\documentclass{article}

\usepackage{graphicx}
\usepackage{hyperref} 
\usepackage{amsmath, amssymb} 
\usepackage{bbold} 
\usepackage{mathtools} 
\usepackage[accepted]{icml2024} 
\usepackage{stfloats} 
\usepackage[nameinlink, capitalise]{cleveref} 

\usepackage{algorithm, algpseudocode}

\definecolor{coolwarm_blue}{rgb}{0.23, 0.299, 0.754}
\definecolor{coolwarm_red}{rgb}{0.706, 0.0156, 0.15}
\definecolor{tab:blue}{rgb}{0.122, 0.467, 0.706}
\definecolor{tab:orange}{rgb}{1.0, 0.498, 0.0549}
\definecolor{tab:green}{rgb}{0.173, 0.627, 0.173}
\definecolor{tab:red}{rgb}{0.839, 0.153, 0.157}
\definecolor{tab:purple}{rgb}{0.58, 0.404, 0.741}
\definecolor{tab:gray}{rgb}{0.502, 0.502, 0.502}
\definecolor{darker_gray}{rgb}{0.35, 0.35, 0.35}
\definecolor{RdYlGn_red}{rgb}{0.65, 0, 0.15}
\definecolor{RdYlGn_green}{rgb}{0, 0.41, 0.22}

\begin{document}

\twocolumn[

\icmltitle{Can a Confident Prior Replace a Cold Posterior?}

\begin{icmlauthorlist}
\icmlauthor{Martin Marek}{ucl}
\icmlauthor{Brooks Paige}{ucl}
\icmlauthor{Pavel Izmailov}{nyu}
\end{icmlauthorlist}
\icmlaffiliation{ucl}{Department of Computer Science, University College London, London, United Kingdom}
\icmlaffiliation{nyu}{Computer Science and Engineering Department, NYU Tandon School of Engineering, New York, USA}
\icmlcorrespondingauthor{Martin Marek}{martin.marek.19@ucl.ac.uk}

\icmlkeywords{ayesian neural networks, Bayesian deep learning, probabilistic deep learning, cold posteriors}

\vskip 0.3in
]

\printAffiliationsAndNotice{}

\begin{abstract}
Benchmark datasets used for image classification tend to have very low levels of label noise. When Bayesian neural networks are trained on these datasets, they often underfit, misrepresenting the aleatoric uncertainty of the data. A common solution is to cool the posterior, which improves fit to the training data but is challenging to interpret from a Bayesian perspective. 
We explore whether posterior tempering can be replaced by a confidence-inducing prior distribution. First, we introduce a \textit{DirClip} prior that is practical to sample and nearly matches the performance of a cold posterior. Second, we introduce a \textit{confidence prior} that directly approximates a cold likelihood in the limit of decreasing temperature but cannot be easily sampled. Lastly, we provide several general insights into confidence-inducing priors, such as when they might diverge and how fine-tuning can mitigate numerical instability. We share our code and weights at \url{https://github.com/martin-marek/dirclip}.
\end{abstract}

\section{Introduction}

\begin{figure}[t]
\begin{center}
\includegraphics[width=\columnwidth]{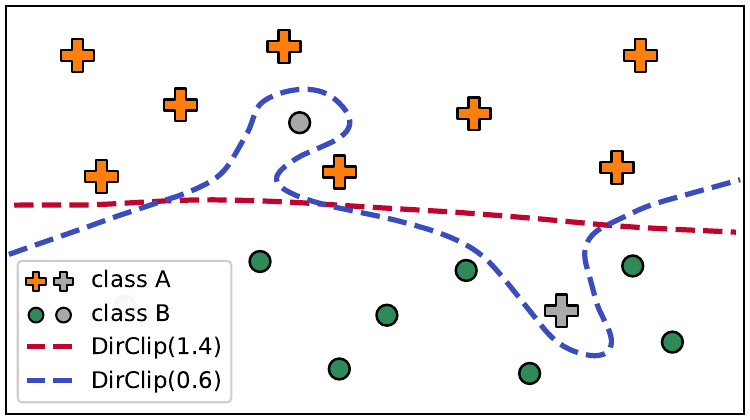}
\vskip -0.1in
\caption{\textbf{Decision boundaries of a Bayesian neural network using the \textit{DirClip} prior.} By varying the concentration parameter of the prior, we can control the model's aleatoric uncertainty, leading to different decision boundaries. The plotted decision boundaries were obtained using Hamiltonian Monte Carlo, using the dataset from Figure 1 of \citet{ndg}.}
\label{fig_dirclip_decision_boundary}
\end{center}
\end{figure}

When performing Bayesian classification, it is important to tune the model's level of aleatoric (data) uncertainty to correctly reflect the noise in the training data. For example, consider the binary classification problem in \Cref{fig_dirclip_decision_boundary}. If we believe that all of the data points were labeled correctly, we would prefer a model that perfectly fits the data, using a {\color{coolwarm_blue}complex} decision boundary. In contrast, if we knew that the data labels were noisy, we might assume that the true decision boundary is actually {\color{coolwarm_red}simpler}, and the two {\color{darker_gray}gray} observations were mislabeled. Both of these decision boundaries provide reasonable descriptions of the data, and we can only choose between them based on our beliefs about the quality of the data labels.

In a regression setting, tuning the model's level of aleatoric uncertainty is a common practice, for example by modifying the kernel of a Gaussian process or by tuning the scale parameter of a Gaussian likelihood \citep{ndg}. In contrast, in a classification setting, we are forced to use the categorical likelihood, which has no tunable parameter to control the level of aleatoric uncertainty. The common solution in practice is to temper the posterior distribution, which softens/sharpens the likelihood. 
However, \citet{cold_posterior} argued that tempering is problematic from a Bayesian perspective as it 1) deviates from the Bayes posterior, and 2) corresponds to a likelihood function that is not a valid distribution over classes. 
In this paper, we aim to show that we can match the results of posterior tempering within the standard Bayesian framework, by introducing a valid prior distribution that directly controls the aleatoric uncertainty of a Bayesian neural network (BNN).

Our paper is heavily inspired by and aims to extend the work of \citet{ndg}. \citeauthor{ndg} have provided a clear mechanistic explanation of how data augmentation causes underfitting, as well as introducing the idea of using a Dirichlet prior to control the aleatoric uncertainty of a BNN. 
We show that the density of the Dirichlet prior that they proposed is unbounded, leading to an improper posterior, which caused the numerical instability in their experiments.
As a result, \citet{ndg} used an approximation, which we show cannot be viewed as purely a prior modification.
In this work, we propose a simple modification of the Dirichlet prior, that fixes the source of the instability, allowing us to match the results of the cold posterior with a pure modification of the prior, without the need for numerical approximations.

We introduce two different and entirely valid prior distributions to control the aleatoric uncertainty of a BNN. First, we introduce the \textit{DirClip} prior---a clipped version of the Dirichlet prior that can be practically sampled and nearly matches the performance of a cold posterior. We explain why clipping (bounding) the prior density is necessary to obtain a valid posterior distribution. We also show that the prior gradient can sometimes dominate the likelihood gradient, leading to unstable training dynamics. We derive exactly when this instability happens and show how initializing the MCMC sampler with a pretrained model can stabilize training.

Second, we introduce a \textit{confidence prior}, which directly enforces low aleatoric uncertainty on the training data. We show that in the limit of decreasing temperature, the confidence prior combined with untempered categorical likelihood converges to a cold likelihood. However, the confidence prior has many local maxima, which makes directly sampling it nearly impossible. While this prior cannot be used practically, it provides a theoretical justification for cold posteriors (which are easy to sample), showing they approximate a valid prior distribution in the limit of decreasing temperature.

\section{Background}

In this section, we introduce the relevant background on Bayesian neural networks and cold posteriors.

\subsection{Bayesian neural networks}
\label{sec_bnn}

Bayesian neural networks are an exciting framework for both understanding and training neural networks that are more reliable \citep{mackay1992bayesian,neal2012bayesian,bdl}. Modern neural networks are typically overparameterized---there are many different sets of model parameters that fit the training data perfectly but disagree on unseen data \citep{mode_connectivity, deep_ens, bdl, underspecification}. As a result, using any single set of parameters to generate predictions is problematic: it ignores our uncertainty over models, leading to over-confident predictions. In a BNN, we consider the full distribution over possible models, leading to both improved accuracy and uncertainty estimation \citep{bnn_posterior}.

A trained BNN is fully defined by its posterior distribution over parameters. Let's denote the model parameters $\boldsymbol{\theta}$, inputs (e.g., images) $\mathbf{X}$ and labels $\mathbf{Y}$. Then the posterior is proportional to a prior times a likelihood:
\begin{align}
\begin{split}
\underbrace{p(\boldsymbol{\theta}|\mathbf{X},\mathbf{Y})}_\text{posterior} \hspace{0.5mm} &\propto \hspace{0.5mm} \underbrace{p(\boldsymbol{\theta}|\mathbf{X})}_\text{prior} \hspace{1mm} \underbrace{p(\mathbf{Y}|\boldsymbol{\theta},\mathbf{X})}_\text{likelihood}.
\end{split}
\label{eq_posterior}
\end{align}
The prior can depend on the input data $\mathbf{X}$ but not on the data labels $\mathbf{Y}$. In practice, the dependence on $\mathbf{X}$ is often ignored, for example by setting a Normal prior over parameters. However, the dependence of the prior on the input data is crucial for functional priors (evaluated on the training data) like the Dirichlet prior \citep{ndg} discussed in \Cref{sec_dirichlet}.

The distribution over the response variable $\mathbf{\tilde{y}}$, given a new observation $\mathbf{\tilde{x}}$ and training data $(\mathbf{X}, \mathbf{Y})$ is given by the \textit{posterior predictive distribution}:
\begin{equation}
p(\mathbf{\tilde{y}} | \mathbf{\tilde{x}}, \mathbf{X}, \mathbf{Y}) = \int p(\mathbf{\tilde{y}} | \mathbf{\tilde{x}}, \boldsymbol{\theta}) p(\boldsymbol{\theta} | \mathbf{X}, \mathbf{Y}) \mathrm{d} \boldsymbol{\theta}.
\label{eq_posterior_predictive_integral}
\end{equation}
Unfortunately, \cref{eq_posterior_predictive_integral} requires integrating over the high-dimensional parameters $\boldsymbol{\theta}$ and does not have a closed-form solution. Hence, we must resort to numerical methods to approximate it. A popular approach is \textit{Monte Carlo simulation}: we express the integral as an expectation of the likelihood over the posterior distribution, draw $N$ samples from the posterior, and compute the empirical mean of the likelihood over these samples:
\begin{align}
\begin{split}
p(\mathbf{\tilde{y}} | \mathbf{\tilde{x}}, \mathbf{X}, \mathbf{Y})
&= \mathbb{E}_{\boldsymbol{\theta} | \mathbf{X}, \mathbf{Y}} [p(\mathbf{\tilde{y}} | \mathbf{\tilde{x}}, \boldsymbol{\theta})] \\
&\approx \sum_{i=1}^{N} \frac{p(\mathbf{\tilde{y}} | \mathbf{\tilde{x}}, \boldsymbol{\theta}_i)}{N}, \boldsymbol{\theta}_i \sim p(\boldsymbol{\theta} | \mathbf{X}, \mathbf{Y}).
\label{eq_sampling_sum}
\end{split}
\end{align}
In this paper, we use Stochastic Gradient Hamiltonian Monte Carlo (SGHMC) to draw the posterior samples and therefore approximate the predictive distribution \citep{sghmc}. We further discuss the SGHMC algorithm in \Cref{appendix_sghmc} and provide our implementation details in \Cref{appendix_implementation_details}.

\subsection{Cold posteriors}
\label{sec_cold_posterior_effects}

A cold posterior is achieved by exponentiating the posterior to $1/T$, where $T<1$. Since the posterior factorizes into a prior and a likelihood, a cold posterior can be seen as a combination of a cold likelihood with a cold prior:
\begin{equation}
p(\boldsymbol{\theta}|\mathbf{X},\mathbf{Y})^{1/T} \propto p(\boldsymbol{\theta}|\mathbf{X})^{1/T} p(\mathbf{Y}|\boldsymbol{\theta},\mathbf{X})^{1/T}.
\label{eq_cold_posterior}
\end{equation}
In practice, a cold posterior is typically sampled by modifying the SGHMC sampling algorithm instead of directly scaling the log-posterior, which could introduce numerical instability (details are provided in \Cref{appendix_sghmc}).
Tempering a Normal prior is equivalent to adjusting its variance by a factor of $T$.
\citet{cold_posterior} showed that the standard posteriors corresponding to $T=1$ lead to poor performance in image classification, while cold posteriors with $T < 1$ provide much better performance. They named this observation the \textit{cold posterior effect}.

In the limit of $T \rightarrow 0$, the cold posterior approaches a deep ensemble \citep{lakshminarayanan2017simple}. 
There are two ways to see this: first, a cold temperature sharpens the posterior. As $T \rightarrow 0$, the exponent approaches $\infty$, so the distribution becomes increasingly sharp, approaching a distribution with point masses located at posterior modes and equal to zero everywhere else (i.e.\ a deep ensemble). At the same time, as $T \rightarrow 0$, the SGHMC noise scale approaches zero, and SGHMC becomes equivalent to SGD with momentum.

\subsection{What is wrong with cold posteriors?}
\label{sec_cold_post_wrong}

In the classification setting, we use a categorical likelihood to represent the predicted probability for each class. A cold posterior corresponds to using a cold categorical likelihood, which is not a valid distribution over classes. A fundamental property of any probability distribution is that the sum of probabilities over all possible outcomes equals one. In cold likelihoods, class probabilities can sum to values significantly less than one, making them invalid as probability distributions \citep{cold_posterior}.

\citet{bdl} argued that posterior tempering can be viewed as adjusting for model misspecification. \citet{ndg} studied cold posteriors in detail, and in particular showed that tempering can counteract the effect of data augmentation. 
In this work, we build on \citet{ndg} and show that it is possible to achieve a similar performance to that of cold posteriors using the standard likelihood function with a modified prior.

\section{Related work}



\textbf{Cold posteriors and data augmentation.}\quad
\citet{bnn_posterior} exactly repeated the experiments of \citet{cold_posterior} but with data augmentation turned off, which entirely eliminated the cold posterior effect. \citet{ndg} have shown that naively implementing data augmentation results in undercounting the training data, which softens the likelihood and directly leads to underfitting. One way to counteract this effect is to use a cold posterior, which sharpens the likelihood and directly compensates for the undercounting.
\citet{principled_da} proposed a non-standard principled version of data augmentation and showed that the cold posterior effect remains in their model, suggesting that other factors can contribute to the cold posterior effect.


\textbf{Label noise.}\quad
Benchmark datasets used for image classification like CIFAR-10 have high label quality \citep{cifar, aleatoric_uncertainty}. Yet when a BNN is trained on CIFAR-10 with data augmentation turned on, the BNN will tend to overestimate the dataset's aleatoric uncertainty, resulting in underfitting \citep{ndg}. 
This holds true across a wide range of popular prior distributions \citep{priors_revisited}. In fact, \citet{cpe_implies_underfitting} have shown that the presence of a cold posterior effect directly implies that the untempered Bayes posterior underfits.

\citet{curation_hypothesis} has argued that the cold posterior effect directly arises from the curation process of benchmark datasets like CIFAR-10. The argument is that these datasets consist entirely of images with clean labels, often requiring the consensus of multiple labelers. They developed a statistical theory which shows that to correctly account for this curation process, BNNs should in fact be using cold posteriors. However, this argument is disputed by \citet{bnn_posterior} who obtained extremely high-fidelity ResNet20 \citep{resnet} posterior samples on CIFAR-10, observing no cold posterior effect.

\textbf{Prior misspecification.}\quad
Intuitively, a Normal prior is not affected by the dataset size but the influence of the likelihood term scales linearly with the dataset size. Therefore, by varying the dataset size, we can change the \textit{relative} influence of the prior. 
\citet{disentangling} trained a BNN on subsets of a dataset of varying size and observed that as the dataset size decreased, the strength of the cold posterior effect increased. This observation implies that the Normal prior is misspecified, especially when the dataset is small or the model is large.
\citet{cold_posterior} and \citet{priors_revisited} also argued that prior misspecification could be one of the key reasons behind the cold posterior effect.

\textbf{Improved prior distributions.}\quad
Based on the prior work discussed above, we conclude that the cold posterior effect has multiple possible causes, although they all seem to stem from underfitting \citep{cpe_implies_underfitting} or overestimating the aleatoric uncertainty of the training data \citep{aleatoric_uncertainty}. We therefore search for a solution in the form of an improved prior distribution, to improve fit to the training data.
 

\citet{priors_revisited} examined the distribution of neural network weights trained using SGD under a uniform prior. They empirically discovered that using heavy-tailed or correlated priors can lead to improved BNN performance. In contrast, the goal of this paper is to theoretically deduce a prior distribution that directly improves the fit on the training data.

\citet{ndg} introduced the idea of using a Dirichlet prior to control the aleatoric uncertainty of a BNN, and therefore remove the need for posterior tempering. Our work is directly inspired by this approach, although we aim to provide a deeper understanding of the Dirichlet prior. We show that the approximation to the Dirichlet prior introduced by \citeauthor{ndg}\ uses a quadratic likelihood term, therefore deviating from a valid distribution over classes. We explain exactly why using an unmodified Dirichlet prior results in divergence and use these insights to develop a valid new prior distribution.

\vspace{-1mm}
\section{Confidence of a Normal prior}
\label{appendix_normal_prior_scale}
\vspace{-0.5mm}

Most prior works studying Bayesian neural networks used isotropic Normal priors \citep{cold_posterior, disentangling, do_bnns_need_to_be_fully_stochastic} or other vague distributions over parameters, such as a Mixture of Gaussians, logistic distribution \citep{bnn_posterior}, Laplace distribution, Student's t-distribution \citep{priors_revisited}, or a correlated Normal distribution \citep{covariate_shift, priors_revisited}. We show that these distributions can have high \textit{prior} confidence, while simultaneously having low \textit{posterior} confidence. This implies that the link between prior and posterior confidence is not trivial, and using a confident prior does not guarantee high posterior confidence.

\begin{figure}[t]
\begin{center}
\includegraphics[width=1\columnwidth]{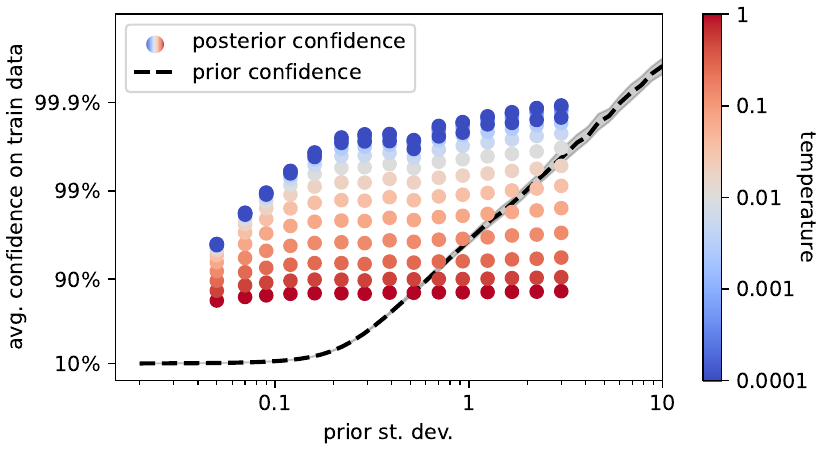}
\vskip -0.1in
\caption{\textbf{Confidence of ResNet20 trained on CIFAR-10 with a Normal prior.} The dashed line shows the average confidence of prior samples as a function of the prior scale (standard deviation). The relationship is one-to-one: the prior scale exactly determines prior confidence. Conversely, the prior scale has almost no effect on posterior confidence---each scatter point corresponds to a single trained model. Here, the intuition that ``prior confidence translates into posterior confidence'' fails. Instead, the posterior confidence depends mostly on the posterior temperature, visualized using the colorbar on the right.
}
\label{fig_normal_conf}
\end{center}
\end{figure}

We performed a simple experiment where we varied the scale of the Normal prior, sampled ResNet20 parameters from the prior, and evaluated its predictions on the training set of CIFAR-10. The dashed line in \Cref{fig_normal_conf} shows the average confidence of prior samples as a function of the prior scale. Intuitively, as the prior scale tends to zero, the model parameters tend to zero, causing logits to approach zero, thereby inducing uniform predictions over class probabilities. Conversely, as the prior scale increases, model parameters grow in magnitude, scaling up the logits. When the scale of logits is large, the absolute differences between the logits grow, leading to high confidence.

Observe in \Cref{fig_normal_conf} that prior confidence does not trivially translate into posterior confidence. Each point corresponds to a single posterior distribution, with color representing the posterior temperature. All {\color{coolwarm_red}untempered $(T=1)$} posteriors have low confidence, irrespective of the prior scale. When a {\color{coolwarm_blue}cold} temperature is used, it is possible to obtain a model that has prior confidence near 10\% (the lowest possible) \textit{and} posterior confidence of 99\%.

These results show that if we want to train a BNN that has high posterior confidence at {\color{coolwarm_red}$T=1$}, it is \textit{not} sufficient to use a prior whose samples are confident. One way to increase the posterior confidence further is to use a prior that will directly assign a high probability density to models with high confidence and a low probability density to models with low confidence. This approach requires a \textit{functional} prior, i.e.\ a prior distribution that is defined over functions (model predictions), rather than model parameters.\footnote{In theory, every distribution over model parameters corresponds to some distribution over functions, and vice versa. We argue that in order to design a confidence-inducing prior distribution, it is \textit{helpful} (although not strictly necessary) to directly think about the prior in function space.}

\begin{figure*}[ht]
\begin{center}
\includegraphics[width=1\textwidth]{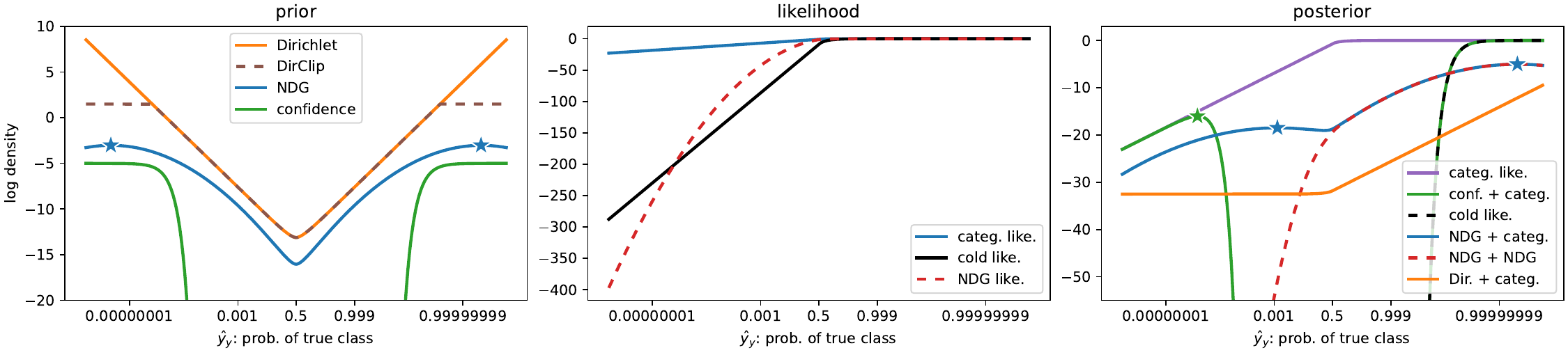}
\vskip -0.1in
\caption{\textbf{Slices of various prior, likelihood, and posterior distributions.} For each distribution, we assume that there are only two classes and we vary the predicted probability of the true class on the x-axis. Since the prior has no notion of the ``true'' class, it is symmetric. Note that the x-axis is non-linear to better show the tail behavior of each distribution. Notably, the NDG prior peaks at a very small (and large) value of predicted probability, which would not be visible on a linear scale. The {\color{tab:blue}blue} and {\color{tab:green}green} stars in the left and right plots show local maxima.}
\label{fig_distribution_slices}
\end{center}
\end{figure*}

\section{Dirichlet prior}
\label{sec_dirichlet}

In a Bayesian classification setting, if we wish to control a model's level of aleatoric uncertainty, the standard approach is to use a Dirichlet prior, which can bias the posterior to either have lower or higher confidence. More formally, the Dirichlet distribution is a distribution over class probabilities. Let's denote a model's predicted class probabilities as $\mathbf{\hat{y}} = (\hat{y}_1, \hat{y}_2 \ldots \hat{y}_{K})$, where $K$ is the number of classes. The Dirichlet prior assigns a probability density to any set of predictions $\mathbf{\hat{y}}$:
\begin{equation}
\log p(\mathbf{\hat{y}}) \stackrel{c}{=} \sum_{k=1}^K (\alpha-1) \log \hat{y}_k.
\label{eq_dir_logpdf}
\end{equation}
The Dirichlet prior is parameterized by a scalar concentration parameter $\alpha$ that biases the predictions toward lower ($\alpha > 1$) or higher ($\alpha < 1$) confidence.\footnote{In general, the concentration parameter is a vector $(\alpha_1, \alpha_2 \ldots \alpha_{K})$ with one component per class. When the components are unequal, the prior has a different bias toward each class. Since we are interested in a symmetric prior, we will only consider the case where each component of this vector is equal.} When $\alpha = 1$, the Dirichlet prior is equal to a uniform prior.

\citet{ndg} have observed that directly using the Dirichlet prior as a prior over parameters $p(\boldsymbol{\theta}) = p(\mathbf{\hat{y}})$ results in divergence. The issue is that the Dirichlet prior is a distribution over model \textit{predictions}, whereas the BNN prior as defined in \cref{eq_posterior} is a distribution over model \textit{parameters}. So how can we translate a distribution over predictions $p(\mathbf{\hat{y}})$ into a distribution over parameters $p(\boldsymbol{\theta})$? In general, this is a difficult problem with no simple solution; we discuss this further in \Cref{appendix_change_of_variables}. In the rest of this paper, we search for a prior that works well when applied directly over model parameters.

To understand why the Dirichlet prior diverges when applied over model parameters, we plot its probability density function (PDF) on the left side of \Cref{fig_distribution_slices}. Notice that the probability density diverges to $\infty$ as $\hat{y}$ approaches either 0 or 1. For a probability distribution to be valid, its PDF needs to integrate to one but the PDF does not necessarily need to be bounded. Indeed, if we treat the Dirichlet prior as a distribution over the domain $\hat{y} \in [0, 1]$, the PDF integrates to one over this domain, and it is, therefore, a valid distribution. The fact that the PDF is unbounded becomes an issue when we treat the PDF as a distribution over parameters because the domain of model parameters is $\boldsymbol{\theta} \in (-\infty, \infty)$. If we attempt to integrate the PDF over the domain of model parameters, we get an integral that diverges, meaning that the distribution is invalid. Even worse, the \textit{posterior} distribution that results from this prior is also invalid and hence impossible to sample. We prove the divergence of the Dirichlet prior more formally in \Cref{appendix_dirichlet_diverges}.

\section{Noisy Dirichlet Gaussian}

To fix the divergence of the Dirichlet prior, \citet{ndg} decided to approximate the \textit{product} (posterior) of the Dirichlet prior with a categorical likelihood. They name this method \textit{Noisy Dirichlet Gaussian (NDG)}. In their experiments, NDG closely matches the performance of a cold posterior.

\begin{figure}[t]
\begin{center}
\includegraphics[width=\columnwidth]{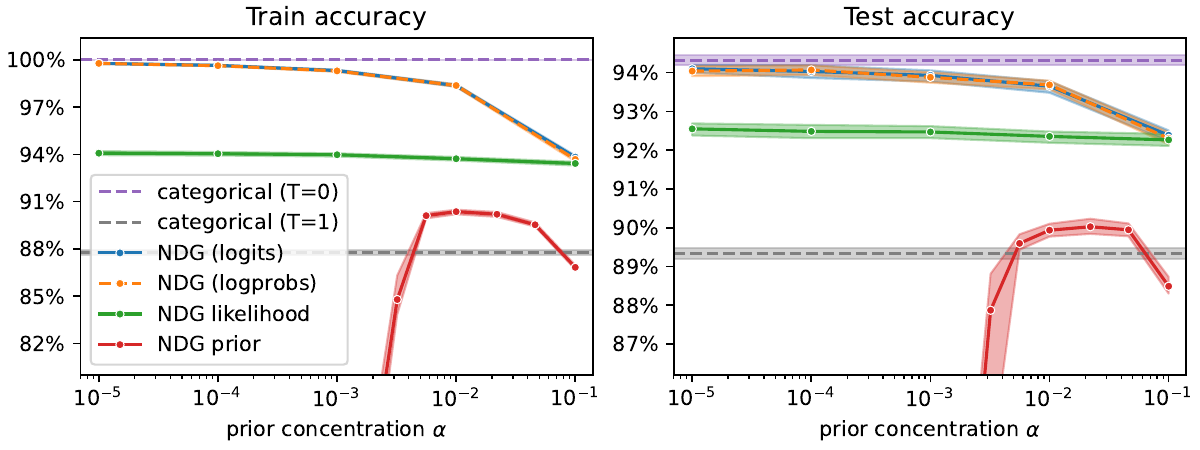}
\vskip -0.1in
\caption{\textbf{Factorized NDG.} This figure shows the accuracy of ResNet20 on CIFAR-10 with data augmentation for various BNN posteriors. Each posterior consists of a $\mathcal{N}(0, 0.1^2)$ prior over model parameters, a (modified) likelihood, and optionally an additional prior term over predictions. The model using the standard {\color{darker_gray}categorical ($T$$=$$1$)} likelihood provides a simple baseline. The NDG posterior models defined over {\color{tab:blue}logits} and {\color{tab:orange}log-probabilities}\footnotemark \ both reach the same test accuracy, on par with a {\color{tab:purple}cold posterior}. In contrast, the NDG {\color{tab:red}prior} and {\color{tab:green}likelihood} on their own do not match the performance of a cold posterior. Note that the training accuracy was evaluated on posterior \textit{samples}, whereas the test accuracy was evaluated on the posterior \textit{ensemble}.}
\vskip -0.2in
\label{fig_ndg_prior_likelihood_accuracy}
\end{center}
\end{figure}

However, we show in \Cref{appendix_ndg} that the NDG posterior approximation corresponds to using a valid prior \textit{combined} with a quadratic likelihood term. This raises the question: does the NDG posterior work so well because of its implied prior distribution or because of the implied quadratic likelihood? To answer this, we perform a simple experiment. First, we train a model that uses the NDG prior together with the categorical likelihood. Second, we train a model that uses the NDG quadratic likelihood. \Cref{fig_ndg_prior_likelihood_accuracy} shows that neither the NDG prior nor the NDG likelihood on their own can match the test accuracy of a cold posterior.

\footnotetext{We discuss the difference between the NDG posterior defined over logits and log-probabilities in \Cref{appendix_ndg}.}

The reason that the NDG prior works at all (unlike a Dirichlet prior) is that its density is bounded. This can be seen in the left plot of \Cref{fig_distribution_slices}: whereas the Dirichlet prior diverges toward each tail of the distribution, the NDG prior density is bounded by a local maximum at each tail of the distribution. If we combine any bounded prior distribution with a Normal prior (with an arbitrarily large scale), we get a valid prior distribution over parameters.\footnotemark

\section{Clipped Dirichlet prior}
\label{sec_dirclip}

\footnotetext{Since the Normal prior is a proper prior, even if we combine it with an improper prior distribution, the joint prior will be proper. In practice, there is little difference between using the NDG prior on its own vs.\ combining it with a large-scale Normal prior; it is simply worth realizing that the latter prior is guaranteed to be proper and therefore result in a proper posterior.}

\begin{figure}[t]
\begin{center}
\includegraphics[width=\columnwidth]{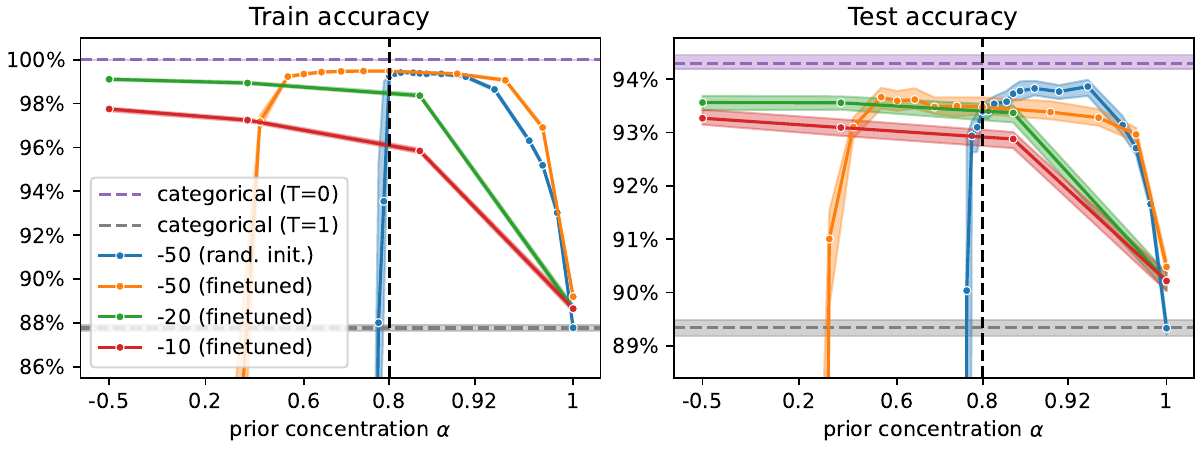}
\vskip -0.1in
\caption{\textbf{DirClip accuracy}. This figure shows the accuracy of ResNet20 on CIFAR-10 with data augmentation for various BNN posteriors. Each solid line corresponds to a different clipping value of the DirClip prior (printed in the legend). The {\color{tab:blue}blue} line shows DirClip posteriors sampled from random initialization; all other DirClip posteriors are fine-tuned from a checkpoint with 100\% training accuracy. Note that the training accuracy was evaluated on posterior \textit{samples}, whereas the test accuracy was evaluated on the posterior \textit{ensemble}.}
\label{fig_dirclip_acc}
\end{center}
\end{figure}

The NDG prior provides an approximation to the Dirichlet prior that fixes its divergence. However, there exists a simpler method to fix the divergence of the Dirichlet prior: we can clip each log-probability at some small value $v$ (for example, $v=-10$):
\begin{equation}
\log p(\mathbf{\hat{y}}) \stackrel{c}{=} \sum_{k=1}^K (\alpha-1) \max(\log\hat{y}_k, v).
\label{eq_dirclip_logpdf}
\end{equation}

Since the log-density of the Dirichlet prior is proportional to the sum of per-class log-probabilities, clipping each log-probability bounds the prior density. We call this prior \textit{DirClip}, as in ``Dirichlet Clipped''. Note that a similar modification of the Dirichlet distribution has been proposed to allow the concentration parameter $\alpha$ to become negative, although it was never applied as a prior over model parameters \citep{clipped_dirichlet}. The DirClip prior is visualized in the left plot of \Cref{fig_distribution_slices}: it is identical to the Dirichlet prior for log-probabilities between the clipping value $v$ and it stays at the clipping value otherwise. We define the DirClip posterior as follows:
\begin{equation}
\underbrace{p(\boldsymbol{\theta}|\mathbf{X},\mathbf{Y})}_{\substack{\text{DirClip}\\\text{posterior}}}
\hspace{0.5mm} \propto \hspace{0.5mm}
\underbrace{p(\boldsymbol{\theta})}_{\substack{\text{Normal}\\\text{prior}}}
\hspace{1mm} \prod_{i=1}^N {\color{darker_gray}
\overbrace{{\color{black} {\underbrace{p(\mathbf{\hat{y}}^{(i)})}_{\substack{\text{DirClip}\\\text{prior}}}} \hspace{1mm}
\underbrace{\hat{y}_y^{(i)}.}_{\substack{\text{categorical}\\\text{likelihood}}}}}^\text{evaluated per observation}} \hspace{-2mm}
\end{equation}

\newpage
\textbf{Results.}\quad We show in \Cref{fig_dirclip_decision_boundary} that the DirClip prior can be used to control the level of aleatoric uncertainty on a toy binary classification dataset. In \Cref{fig_dirclip_acc}, we show that the DirClip prior can be used to control the training accuracy of a ResNet20 trained on CIFAR-10. When $\alpha=1$, the DirClip prior is equivalent to a uniform prior, and the model underfits. By increasing the prior concentration (reducing $\alpha$), we can increase training accuracy from 88\% to 99\% and test accuracy from 89\% to almost 94\%. However, notice in the figure that the model trained from {\color{tab:blue}random initialization} only converges when $\alpha > 0.8$. When $\alpha<0.8$, the accuracy suddenly drops all the way down to 10\% (corresponding to random guessing). This sudden drop in accuracy is not a fundamental property of the DirClip prior; instead, it can be entirely attributed to the challenges of gradient-based sampling algorithms.

\subsection{Training stability}

\begin{figure}[t]
\begin{center}
\includegraphics[width=\columnwidth]{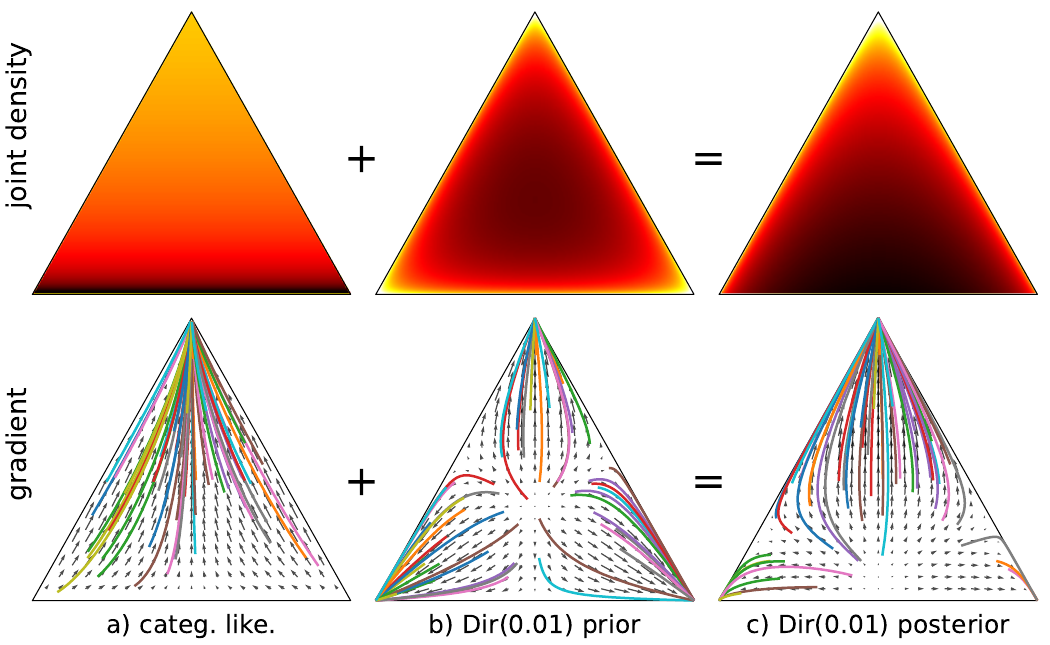}
\vskip -0.08in
\caption{\textbf{Top:} probability density of a) categorical likelihood, b) Dirichlet prior, c) Dirichlet prior combined with categorical likelihood. Each distribution is defined over three classes. The top corner corresponds to the correct class and the two bottom corners correspond to the two other classes. \textbf{Bottom:} vector field shows the direction and magnitude of gradients computed in logit space and reprojected back to the probability simplex. The colored lines show the trajectories of 50 randomly sampled particles under this gradient field.} 
\label{fig_dirichlet_gradient}
\end{center}
\end{figure}

\Cref{fig_dirichlet_gradient} shows the paths that 50 randomly sampled particles would take along the domain of various probability distributions under gradient descent in logit space. Notice that the gradient of the categorical likelihood always points toward the top corner (correct class). In contrast, the Dirichlet prior has no notion of a ``correct'' class, and its gradient simply points toward whichever corner (class) is the closest. When the Dirichlet prior is combined with the categorical likelihood, the combined gradient does not necessarily point toward the correct class. For example, in \hyperref[fig_dirichlet_gradient]{Figure~\ref*{fig_dirichlet_gradient}c}, around 30\% of uniformly-sampled particles would converge to either of the bottom corners under gradient descent. However, if we directly sampled the Dirichlet posterior, only 1\% of the samples would appear in the bottom half of the plot. Therefore, this behavior of particles converging to the wrong class is purely an optimization problem, rather than a statistical property of the Dirichlet distribution.

The same behavior translates to a Bayesian neural network optimized (sampled) using gradient-based methods. If the neural network predicts a wrong class at initialization, optimizing the parameters of the neural network may further \textit{increase} the model's predicted probability for that class. The model does not necessarily converge to predicting the correct class.

\begin{figure}[t]
\begin{center}
\includegraphics[width=0.8\columnwidth]{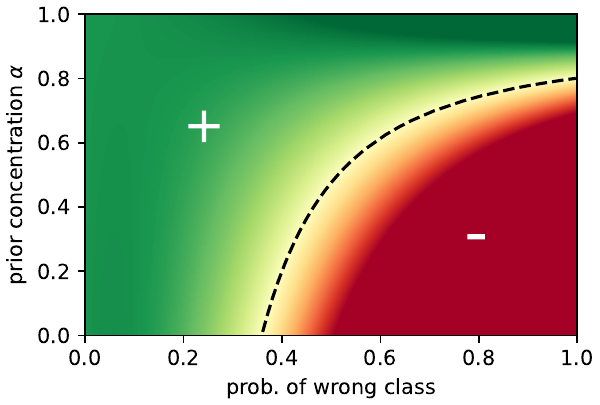}
\vskip -0.1in
\caption{\textit{Does a gradient step on the Dirichlet posterior increase the probability of the correct class?} Assume that there are 10 classes. The x-axis shows the probability of a \textit{wrong} class; the other 9 classes (one of which is the correct class) have the same probability. The image color shows the change in the probability of the correct class under a gradient step. {\color{RdYlGn_green}Green} means that the probability of the correct class will increase; {\color{RdYlGn_red}red} means that it will decrease. Notice that for $\alpha > 0.8$, the probability of the correct class always increases. For $\alpha < 0.8$, the probability of the correct class may increase or decrease, depending on the x-axis value.}
\label{fig_dirichlet_phase_diagram}
\end{center}
\end{figure}

In the case of the Dirichlet posterior, we can analytically derive whether a gradient step will increase or decrease the probability of the true class (full derivation is provided in \Cref{appendix_dirichlet_gradient_direction}). Intuitively, if a model's prediction is initially close to the true class, then gradient descent will increase the probability of that class. However, when the prediction is initially close to a wrong class, the gradient \textit{may} point toward the wrong class. \Cref{fig_dirichlet_phase_diagram} visualizes this behavior using a phase diagram. If a neural network is randomly initialized and $\alpha < 0.8$, most of its predictions will fall inside the {\color{RdYlGn_red}diverging} phase, meaning those predictions may not converge toward the true class. There are two ways to fix this: 1) use $\alpha > 0.8$; or 2) initialize the neural network such that all of its predictions fall inside the {\color{RdYlGn_green}converging} phase. In practical terms, this means fine-tuning the neural network from a checkpoint that has 100\% training accuracy. In \Cref{fig_dirclip_acc}, we show that the fine-tuning strategy works for $\alpha < 0.8$.\footnote{fine-tuning is necessary but not sufficient for convergence with $\alpha < 0.8$. In particular, we found that smaller values of $\alpha$ require increasingly smaller learning rates. The stability phase diagram in \Cref{fig_dirichlet_phase_diagram} only holds for an infinitely small learning rate.} For further discussion of the DirClip prior, please see \Cref{appendix_dirclip}. 

\section{Confidence prior}
\label{sec_conf_prior}

\begin{figure}[t]
\begin{center}
\includegraphics[width=1\columnwidth]{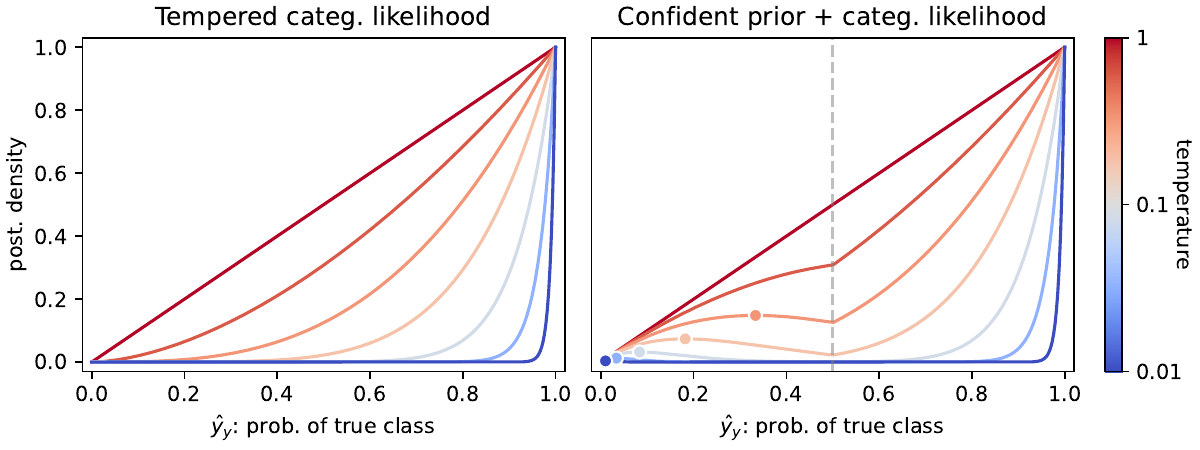}
\vskip -0.1in
\caption{\textbf{Left:} Density of the categorical likelihood tempered to various temperatures (shown by the colorbar on the right). A {\color{coolwarm_blue}cold} likelihood ($T < 1$) is ``sharp'' and penalizes wrong predictions heavily. The {\color{coolwarm_red}red} line shows the untempered likelihood ($T = 1$). \textbf{Right:} An upper bound on the product of a confident prior (parameterized using temperature) with the untempered categorical likelihood. Note that for $\hat{y}_y > 0.5$, the two densities are identical.}
\label{fig_conf_pdf}
\end{center}
\end{figure}

While the DirClip prior \textit{can} successfully control a model's level of aleatoric uncertainty, it achieves this in a very indirect way. To see this, consider the probability density that the DirClip$(0.01)$ prior assigns to the following two predictions:
\vspace{-1mm}
\begin{align}
\begin{split}
p((99\%, 0.5\%, 0.5\%)) &\approx 3.6 \\
p((49.99999\%, 49.99999\%, 0.00002\%)) &\approx 1102.
\end{split}
\end{align}
The probability density assigned to the second prediction is roughly \textit{300-times} higher even though it only has half the confidence of the first prediction ($\sim$50\% vs.\ 99\%). The issue is that the DirClip prior (and the Dirichlet distribution in general) assigns a high density to predictions with \textit{small probabilities} rather than predictions with high confidence. This can be seen directly by looking at the Dirichlet PDF in \cref{eq_dir_logpdf}: when $\alpha < 1$, the density is high when any of the log-probabilities is extremely small.
In particular, high density of the Dirichlet distribution does not directly imply high confidence.

Instead of using a prior that encourages small log-probabilities, we can design a prior that directly enforces high prediction confidence. We can simply set the density of the prior proportional to the confidence (i.e.\ the maximum predicted probability):
\begin{equation}
p(\boldsymbol{\hat{y}}) \propto \left(\max_k \hat{y}_k \right)^{1/T-1}.
\label{eq_conf_prior}
\end{equation}
We intentionally parameterized this ``confidence prior'' using $T$, so that when the predicted class matches the true class (which is guaranteed to be true when $\hat{y}_y \geq 0.5$), the density of the confidence prior combined with categorical likelihood is equal to the density of a cold categorical likelihood:
\begin{equation}
\underset{k}{\mathrm{argmax}} \hspace{0.8mm} \hat{y}_k = y \implies  p(\boldsymbol{\hat{y}}) \cdot \hat{y}_y = \hat{y}_y^{1/T}.
\label{eq_conf_equals_cold_likelihood}
\end{equation}

On the left side of \Cref{fig_conf_pdf}, we plot the density of tempered categorical likelihood, which also provides a lower bound on the product of the confidence prior with untempered categorical likelihood. On the right side, we plot the upper bound. Notice that as the temperature approaches {\color{coolwarm_blue}zero}, both densities converge to the Dirac delta measure concentrated at 1. We prove this convergence in \Cref{appendix_confidence_prior}.

Recall from \cref{eq_cold_posterior} that when a Normal prior and categorical likelihood is used, a cold posterior is equivalent to rescaling the prior and tempering the likelihood. However, it is also possible to approximate the cold posterior with a ``confidence posterior''. The ``confidence posterior'' consists of a rescaled Normal prior, confidence prior, and \textit{untempered} categorical likelihood:
\vspace{-1mm}
\begin{equation}
\underbrace{p(\boldsymbol{\theta}|\mathbf{X},\mathbf{Y})^\frac{1}{T}}_\text{cold posterior}
\approx \underbrace{p(\boldsymbol{\theta}|T\sigma^2)}_{\substack{\text{rescaled}\\\text{Normal prior}}} \prod_{i=1}^N {\color{coolwarm_blue} \overbrace{ {\color{black} {\underbrace{\left(\max_k \hat{y}_k^{(i)} \right)}_\text{confidence prior}}^{\hspace{-1mm} \frac{1}{T}-1} \hspace{-2mm} \underbrace{\hat{y}_y^{(i)}.}_{\substack{\text{categorical}\\\text{likelihood}}}}}^\text{approximates cold likelihood}} \hspace{-2mm}
\end{equation}

In fact, if the posterior mode reaches 100\% accuracy, then the confidence posterior and cold posterior have the same Hessian, implying that the two distributions have an identical Laplace approximation. This follows directly from \cref{eq_conf_equals_cold_likelihood}.

\begin{figure*}[t]
\begin{center}
\includegraphics[width=1\textwidth]{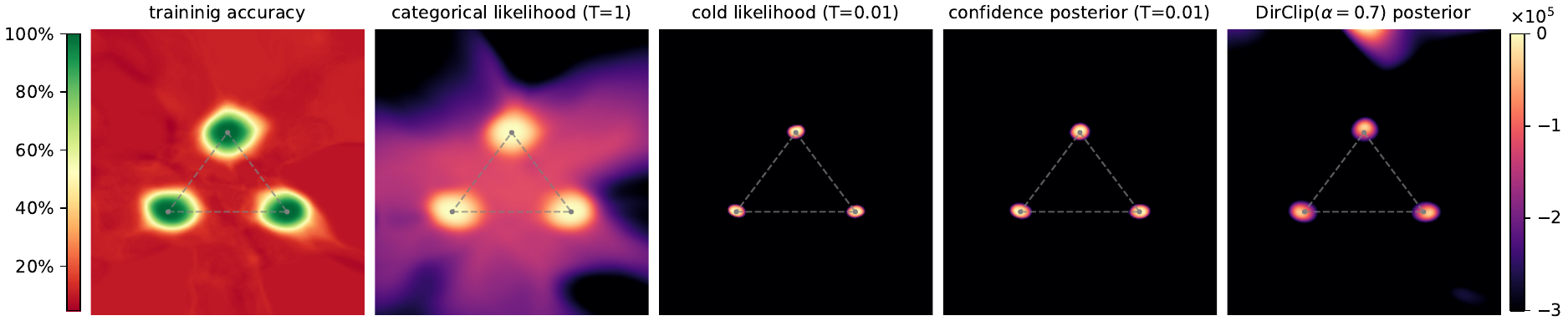}
\vskip -0.1in
\caption{\textbf{Posterior landscapes.} Each plot shows a gray triangle whose vertices correspond to different ResNet20 models trained on CIFAR-10 using SGD. We evaluate the training accuracy and various distributions along the plane defined by the parameters of the three trained models. The colorbar on the left shows the accuracy scale and the colorbar on the right shows the distribution log-density.} 
\label{fig_loss_landscapes_5x}
\end{center}
\end{figure*}

In \Cref{fig_loss_landscapes_5x}, we visualize various posterior distributions. Notice that the \textit{categorical likelihood} is relatively diffuse, assigning high probability density to models that deviate from the SGD solutions. In contrast, both the \textit{cold likelihood} and the \textit{confidence posterior} are very ``sharp''---they are concentrated closely around the three trained models. Lastly, the DirClip posterior shows a surprising behavior, assigning the highest probability density to a model that only has a 10\% training accuracy. The reason is that the top region of the plot happens to minimize the predicted log-probabilities and therefore maximize the DirClip prior density.

Unfortunately, it is not possible to sample the confidence posterior using SGHMC or related optimization methods. The issue is that, unlike a cold likelihood, the confidence prior combined with categorical likelihood is riddled with local maxima. In \Cref{fig_conf_pdf}, each dot in the right plot shows a local maximum, using a linear y-axis scale. While these maxima may not seem too steep, they get steeper with decreasing temperature. The right plot in \Cref{fig_distribution_slices} shows the same local maximum on a logarithmic scale for a lower temperature parameter $\left(T=3\cdot10^{-7}\right)$. In theory, if we were sampling points uniformly under the ``confidence posterior'' curve in \Cref{fig_distribution_slices}, more than 99.9999\% of the sampled points would end up on the right. However, local optimization methods such as SGD and SGHMC might get stuck in the local minimum on the left. Depending on how these local optimization algorithms are initialized, up to 50\% of the sampled points might end up on the left.\footnote{Sampling algorithms like HMC generate unbiased but correlated posterior samples. In theory, the posterior approximation gets more accurate the longer we run the sampling algorithm. In practice, however, it might be infeasible to run the sampling algorithm for ``long enough''.}

\section{Discussion}

When a BNN is trained on a curated dataset like CIFAR-10, it might overestimate the aleatoric uncertainty of the data, therefore underfitting. The standard solution is to cool the posterior distribution, which improves fit to the training data. However, tempering might seem problematic from a theoretical point of view, as it deviates from the Bayes posterior and corresponds to an invalid distribution over classes. We show that we can achieve similar results to posterior tempering purely within the standard Bayesian framework, through the combination of the standard categorical likelihood with a confidence-inducing prior distribution.




\textbf{\textit{DirClip} prior controls aleatoric uncertainty}. By clipping the density of the Dirichlet prior, we fixed its divergence, leading to a valid prior distribution that controls the level of aleatoric uncertainty. When applied to ResNet20 on CIFAR-10, the DirClip prior nearly matches the accuracy of a cold posterior, without the need for any tempering. We also examined the gradient field of the prior, explaining why it is difficult to sample for a small concentration parameter.

\textbf{\textit{Confidence posterior} approximates a cold posterior.} By observing that the Dirichlet prior encourages small probabilities (rather than high confidence), we were able to design a prior that directly enforces high confidence. When this prior is combined with a categorical likelihood, as $T \rightarrow 0$, their product converges to a cold likelihood. This allows us to see the cold posterior of a BNN from a new perspective: not as corresponding to a cold likelihood, but instead as approximating the confidence prior combined with the \textit{untempered} categorical likelihood.

\textbf{Summary.} We acknowledge that cold posteriors are significantly easier to sample than confidence-inducing priors, and therefore remain the practical solution when the model is misspecified \citep{bdl}. The goal of this paper was to show that 1) tempering is not \textit{necessary} to achieve high accuracy on CIFAR-10; and 2) cold posteriors can be seen as approximating a valid prior distribution.


\section*{Impact statement}

This paper presents work whose goal is to advance Bayesian deep learning research, but at substantial computational expense. We estimate that our experiments used over 2,000 kWh of electricity. Broader adoption of similar techniques could carry a large carbon footprint. While we believe there is merit in trading computation for statistical fidelity, we encourage future work to prioritize energy-efficient implementations of similar methods.

\section*{Acknowledgements}
This research was supported by Google's TPU Research Cloud (TRC) program: \url{https://sites.research.google/trc/}.

\bibliography{paper.bib}

\begin{thebibliography}{27}
\providecommand{\natexlab}[1]{#1}
\providecommand{\url}[1]{\texttt{#1}}
\expandafter\ifx\csname urlstyle\endcsname\relax
  \providecommand{\doi}[1]{doi: #1}\else
  \providecommand{\doi}{doi: \begingroup \urlstyle{rm}\Url}\fi

\bibitem[Adlam et~al.(2020)Adlam, Snoek, and Smith]{aleatoric_uncertainty}
Adlam, B., Snoek, J., and Smith, S.~L.
\newblock Cold posteriors and aleatoric uncertainty.
\newblock \emph{arXiv preprint arXiv:2008.00029}, 2020.

\bibitem[Aitchison(2021)]{curation_hypothesis}
Aitchison, L.
\newblock A statistical theory of cold posteriors in deep neural networks.
\newblock In \emph{International Conference on Learning Representations}, 2021.
\newblock URL \url{https://openreview.net/forum?id=Rd138pWXMvG}.

\bibitem[Bradbury et~al.(2018)Bradbury, Frostig, Hawkins, Johnson, Leary, Maclaurin, Necula, Paszke, Vander{P}las, Wanderman-{M}ilne, and Zhang]{jax}
Bradbury, J., Frostig, R., Hawkins, P., Johnson, M.~J., Leary, C., Maclaurin, D., Necula, G., Paszke, A., Vander{P}las, J., Wanderman-{M}ilne, S., and Zhang, Q.
\newblock {JAX}: composable transformations of {P}ython+{N}um{P}y programs, 2018.
\newblock URL \url{http://github.com/google/jax}.

\bibitem[Brooks \& Gelman(1998)Brooks and Gelman]{r_hat}
Brooks, S.~P. and Gelman, A.
\newblock General methods for monitoring convergence of iterative simulations.
\newblock \emph{Journal of computational and graphical statistics}, 7\penalty0 (4):\penalty0 434--455, 1998.

\bibitem[Chen et~al.(2014)Chen, Fox, and Guestrin]{sghmc}
Chen, T., Fox, E., and Guestrin, C.
\newblock Stochastic gradient hamiltonian monte carlo.
\newblock In \emph{International conference on machine learning}, pp.\  1683--1691. PMLR, 2014.

\bibitem[D'Amour et~al.(2020)D'Amour, Heller, Moldovan, Adlam, Alipanahi, Beutel, Chen, Deaton, Eisenstein, Hoffman, et~al.]{underspecification}
D'Amour, A., Heller, K., Moldovan, D., Adlam, B., Alipanahi, B., Beutel, A., Chen, C., Deaton, J., Eisenstein, J., Hoffman, M.~D., et~al.
\newblock Underspecification presents challenges for credibility in modern machine learning.
\newblock \emph{arXiv preprint arXiv:2011.03395}, 2020.

\bibitem[Fort et~al.(2019)Fort, Hu, and Lakshminarayanan]{deep_ens}
Fort, S., Hu, H., and Lakshminarayanan, B.
\newblock Deep ensembles: A loss landscape perspective.
\newblock \emph{arXiv preprint arXiv:1912.02757}, 2019.

\bibitem[Fortuin et~al.(2021)Fortuin, Garriga-Alonso, Wenzel, Ratsch, Turner, van~der Wilk, and Aitchison]{priors_revisited}
Fortuin, V., Garriga-Alonso, A., Wenzel, F., Ratsch, G., Turner, R.~E., van~der Wilk, M., and Aitchison, L.
\newblock Bayesian neural network priors revisited.
\newblock In \emph{Third Symposium on Advances in Approximate Bayesian Inference}, 2021.
\newblock URL \url{https://openreview.net/forum?id=xaqKWHcoOGP}.

\bibitem[Garipov et~al.(2018)Garipov, Izmailov, Podoprikhin, Vetrov, and Wilson]{mode_connectivity}
Garipov, T., Izmailov, P., Podoprikhin, D., Vetrov, D.~P., and Wilson, A.~G.
\newblock Loss surfaces, mode connectivity, and fast ensembling of dnns.
\newblock \emph{Advances in neural information processing systems}, 31, 2018.

\bibitem[He et~al.(2016)He, Zhang, Ren, and Sun]{resnet}
He, K., Zhang, X., Ren, S., and Sun, J.
\newblock Deep residual learning for image recognition.
\newblock In \emph{Proceedings of the IEEE conference on computer vision and pattern recognition}, pp.\  770--778, 2016.

\bibitem[Izmailov et~al.(2021{\natexlab{a}})Izmailov, Nicholson, Lotfi, and Wilson]{covariate_shift}
Izmailov, P., Nicholson, P., Lotfi, S., and Wilson, A.~G.
\newblock Dangers of bayesian model averaging under covariate shift.
\newblock \emph{Advances in Neural Information Processing Systems}, 34:\penalty0 3309--3322, 2021{\natexlab{a}}.

\bibitem[Izmailov et~al.(2021{\natexlab{b}})Izmailov, Vikram, Hoffman, and Wilson]{bnn_posterior}
Izmailov, P., Vikram, S., Hoffman, M.~D., and Wilson, A. G.~G.
\newblock What are bayesian neural network posteriors really like?
\newblock In \emph{International Conference on Machine Learning}, pp.\  4629--4640. PMLR, 2021{\natexlab{b}}.

\bibitem[Kapoor et~al.(2022)Kapoor, Maddox, Izmailov, and Wilson]{ndg}
Kapoor, S., Maddox, W., Izmailov, P., and Wilson, A.~G.
\newblock On uncertainty, tempering, and data augmentation in bayesian classification.
\newblock In Oh, A.~H., Agarwal, A., Belgrave, D., and Cho, K. (eds.), \emph{Advances in Neural Information Processing Systems}, 2022.
\newblock URL \url{https://openreview.net/forum?id=pBJe5yu41Pq}.

\bibitem[Krizhevsky et~al.(2009)Krizhevsky, Hinton, et~al.]{cifar}
Krizhevsky, A., Hinton, G., et~al.
\newblock Learning multiple layers of features from tiny images.
\newblock 2009.

\bibitem[Lakshminarayanan et~al.(2017)Lakshminarayanan, Pritzel, and Blundell]{lakshminarayanan2017simple}
Lakshminarayanan, B., Pritzel, A., and Blundell, C.
\newblock Simple and scalable predictive uncertainty estimation using deep ensembles.
\newblock \emph{Advances in neural information processing systems}, 30, 2017.

\bibitem[Mackay(1992)]{mackay1992bayesian}
Mackay, D. J.~C.
\newblock \emph{Bayesian methods for adaptive models}.
\newblock California Institute of Technology, 1992.

\bibitem[Nabarro et~al.(2022)Nabarro, Ganev, Garriga-Alonso, Fortuin, van~der Wilk, and Aitchison]{principled_da}
Nabarro, S., Ganev, S., Garriga-Alonso, A., Fortuin, V., van~der Wilk, M., and Aitchison, L.
\newblock Data augmentation in bayesian neural networks and the cold posterior effect.
\newblock In \emph{Uncertainty in Artificial Intelligence}, pp.\  1434--1444. PMLR, 2022.

\bibitem[Neal(2012)]{neal2012bayesian}
Neal, R.~M.
\newblock \emph{Bayesian learning for neural networks}, volume 118.
\newblock Springer Science \& Business Media, 2012.

\bibitem[Neal et~al.(2011)]{hmc}
Neal, R.~M. et~al.
\newblock Mcmc using hamiltonian dynamics.
\newblock \emph{Handbook of markov chain monte carlo}, 2\penalty0 (11):\penalty0 2, 2011.

\bibitem[Noci et~al.(2021)Noci, Roth, Bachmann, Nowozin, and Hofmann]{disentangling}
Noci, L., Roth, K., Bachmann, G., Nowozin, S., and Hofmann, T.
\newblock Disentangling the roles of curation, data-augmentation and the prior in the cold posterior effect.
\newblock \emph{Advances in Neural Information Processing Systems}, 34:\penalty0 12738--12748, 2021.

\bibitem[Qiu et~al.(2023)Qiu, Rudner, Kapoor, and Wilson]{fs_map}
Qiu, S., Rudner, T. G.~J., Kapoor, S., and Wilson, A.~G.
\newblock Should we learn most likely functions or parameters?
\newblock In \emph{Thirty-seventh Conference on Neural Information Processing Systems}, 2023.
\newblock URL \url{https://openreview.net/forum?id=9EndFTDiqh}.

\bibitem[Sharma et~al.(2023)Sharma, Farquhar, Nalisnick, and Rainforth]{do_bnns_need_to_be_fully_stochastic}
Sharma, M., Farquhar, S., Nalisnick, E., and Rainforth, T.
\newblock Do bayesian neural networks need to be fully stochastic?
\newblock In Ruiz, F., Dy, J., and van~de Meent, J.-W. (eds.), \emph{Proceedings of The 26th International Conference on Artificial Intelligence and Statistics}, volume 206 of \emph{Proceedings of Machine Learning Research}, pp.\  7694--7722. PMLR, 25--27 Apr 2023.
\newblock URL \url{https://proceedings.mlr.press/v206/sharma23a.html}.

\bibitem[Tu(2016)]{clipped_dirichlet}
Tu, K.
\newblock Modified dirichlet distribution: Allowing negative parameters to induce stronger sparsity.
\newblock In \emph{Proceedings of the 2016 Conference on Empirical Methods in Natural Language Processing}, pp.\  1986--1991, 2016.

\bibitem[Welling \& Teh(2011)Welling and Teh]{sgld}
Welling, M. and Teh, Y.~W.
\newblock Bayesian learning via stochastic gradient langevin dynamics.
\newblock In \emph{Proceedings of the 28th international conference on machine learning (ICML-11)}, pp.\  681--688, 2011.

\bibitem[Wenzel et~al.(2020)Wenzel, Roth, Veeling, Swiatkowski, Tran, Mandt, Snoek, Salimans, Jenatton, and Nowozin]{cold_posterior}
Wenzel, F., Roth, K., Veeling, B., Swiatkowski, J., Tran, L., Mandt, S., Snoek, J., Salimans, T., Jenatton, R., and Nowozin, S.
\newblock How good is the {B}ayes posterior in deep neural networks really?
\newblock In III, H.~D. and Singh, A. (eds.), \emph{Proceedings of the 37th International Conference on Machine Learning}, volume 119 of \emph{Proceedings of Machine Learning Research}, pp.\  10248--10259. PMLR, 13--18 Jul 2020.
\newblock URL \url{https://proceedings.mlr.press/v119/wenzel20a.html}.

\bibitem[Wilson \& Izmailov(2020)Wilson and Izmailov]{bdl}
Wilson, A.~G. and Izmailov, P.
\newblock Bayesian deep learning and a probabilistic perspective of generalization.
\newblock \emph{Advances in neural information processing systems}, 33:\penalty0 4697--4708, 2020.

\bibitem[Zhang et~al.(2023)Zhang, Wu, Ortega, and Masegosa]{cpe_implies_underfitting}
Zhang, Y., Wu, Y.-S., Ortega, L.~A., and Masegosa, A.~R.
\newblock If there is no underfitting, there is no cold posterior effect.
\newblock \emph{arXiv preprint arXiv:2310.01189}, 2023.

\end{thebibliography}
\bibliographystyle{icml2024}

\clearpage
\appendix

\begin{figure*}[ht]
\begin{center}
\includegraphics[width=\textwidth]{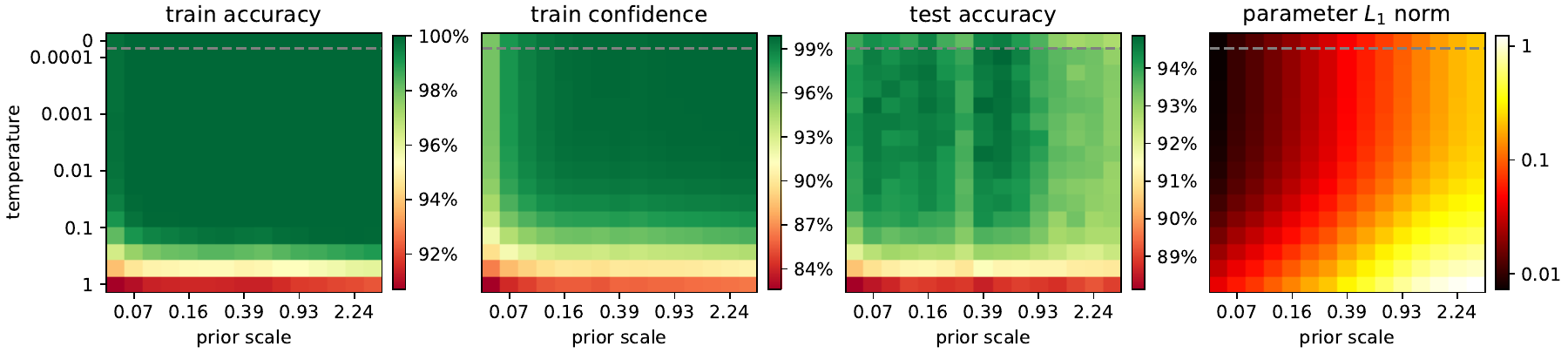}
\vskip -0.1in
\caption{\textbf{Effect of posterior temperature and Normal prior scale on a Bayesian neural network.} The model is a ResNet20 trained on CIFAR-10 with data augmentation turned on. When $T=1$, the augmented data is undercounted, resulting in underfitting (low training accuracy). As temperature decreases, the fit to training data improves, increasing test accuracy. At the same time, as temperature decreases, the norm of the model parameters decreases, but only up to a point. With a decreasing temperature, the cold posterior approaches a deep ensemble, which is obtained by setting the temperature to exactly zero.}
\label{fig_temp_grid}
\end{center}
\end{figure*}

\section*{Appendix outline}
\begin{itemize}
    \item In \cref{appedix_model_comparison}, we visualize the effect of posterior tempering and compare the accuracy and likelihood achieved by all models discussed in this paper. 
    \item In \cref{appendix_dirclip}, we show that the DirClip prior reaches its clipping value, which results in low test likelihood when the posterior distribution is approximated using a small number of samples.
    \item In \cref{appendix_ndg}, we factorize the NDG posterior into a prior and a likelihood.
    \item In \cref{appendix_confidence_prior}, we prove that the confidence prior converges to a cold likelihood.
    \item In \cref{appendix_priors_over_outputs}, we explain why correctly sampling from a prior over model outputs requires a ``change of variables'' term; we prove that the Dirichlet prior diverges if this term is omitted; and we prove that the DirClip prior is a valid distribution.
    \item In \cref{appendix_dirichlet_gradient_direction}, we derive the direction of a gradient step along the Dirichlet posterior and show when it increases the probability of the observed class.
    \item In \cref{appendix_implementation_details}, we provide implementation details for all of our experiments.
    \item In \cref{appendix_mcmc}, we further describe the HMC and SGHMC algorithms.
\end{itemize}

\section{Model comparison}
\label{appedix_model_comparison}

To visualize the effect of posterior tempering on the training accuracy, test accuracy, and parameter norm, we sampled 16 different posterior temperatures across 15 different prior scales---the results are shown in \Cref{fig_temp_grid}.

\begin{figure*}[ht]
\begin{center}
\includegraphics[width=0.85\textwidth]{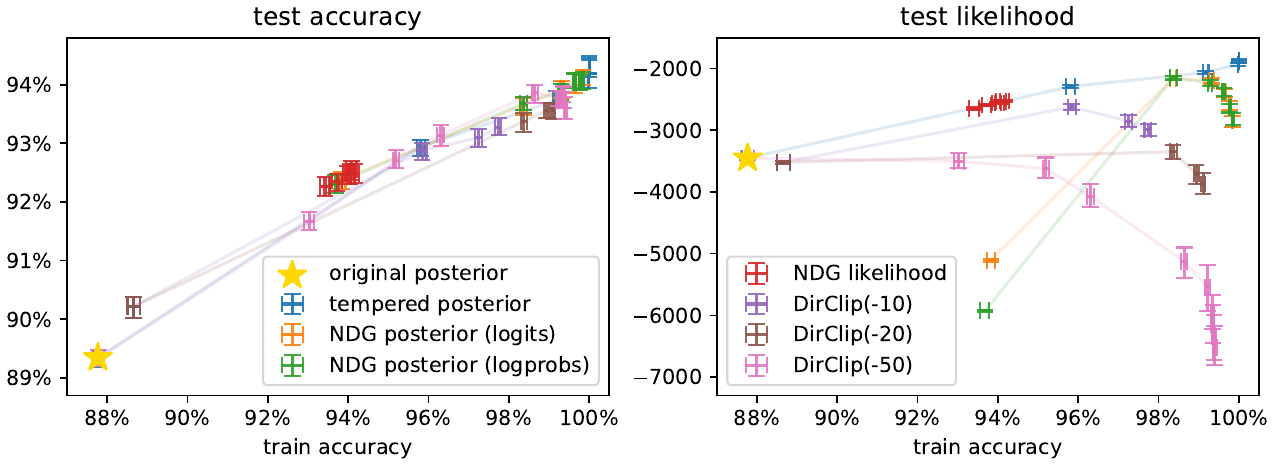}
\vskip -0.1in
\caption{\textbf{Comparison of various methods for controlling aleatoric uncertainty.} The model is a ResNet20 trained on CIFAR-10 with data augmentation turned on. The \textit{original posterior} consists of a $\mathcal{N}(0, 0.1^2)$ prior over model parameters and the categorical likelihood; the \textit{tempered posterior} uses the same prior and likelihood, but the posterior temperature is varied. The \textit{NDG} models use a $\mathcal{N}(0, 0.1^2)$ prior over parameters, combined with either the full NDG posterior or only the quadratic NDG likelihood on its own. Lastly, the DirClip models use varying concentration parameters $\alpha$ with the clipping value shown in the plot legend. }
\label{fig_model_comparison}
\end{center}
\end{figure*}

In \Cref{fig_model_comparison}, we compare the training accuracy, test accuracy, and test likelihood of various posterior distributions. Observe in the left plot that the training accuracy almost perfectly predicts the test accuracy. A possible explanation behind this effect is that most of the posteriors underfit the training data, therefore overestimating the aleatoric uncertainty in the labels \citep{aleatoric_uncertainty}. In order to correctly represent the aleatoric uncertainty in the data labels, the model needs to reach near 100\% training accuracy. However, the behavior of the test likelihood is more complex---we discuss this in \Cref{appendix_dirclip_low_likelihood}.

\section{Further discussion of DirClip prior}
\label{appendix_dirclip}

\subsection{Clipping value is reached}

\begin{figure}[t]
\begin{center}
\includegraphics[width=0.9\columnwidth]{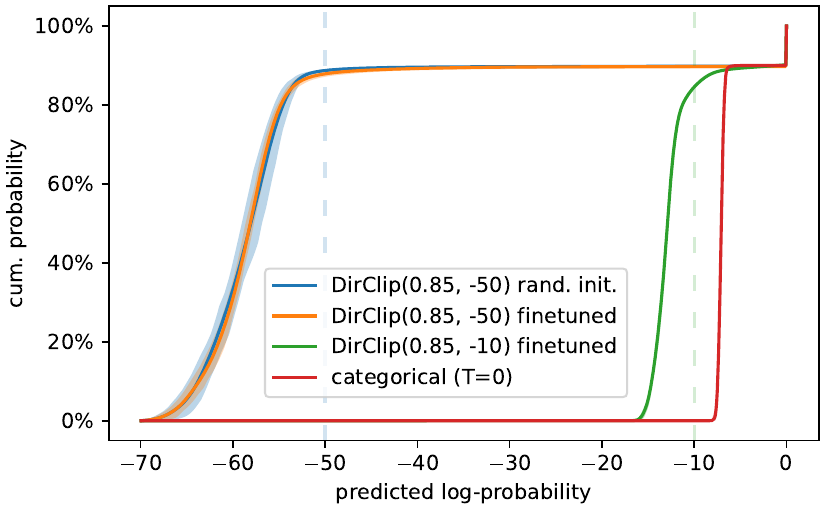}
\vskip -0.1in
\caption{\textbf{CDF of predicted log-probabilities on training data.} We sampled various ResNet20 posteriors on CIFAR-10. Afterwards, we independently evaluated the predictions of each posterior sample on the training data, concatenating all predicted log-probabilities across classes and posterior samples.}
\label{fig_logprob_cdf}
\end{center}
\end{figure}

In \Cref{fig_dirclip_acc}, we show that the DirClip prior achieves higher training accuracy when the clipping value $-50$ is used, compared to the clipping value $-10$. This might seem surprising because a log-probability of $-10$ is already a very small value. The reason the clipping value affects the prior behavior is that the clipping value \textit{is} reached on most predictions. Observe in \Cref{fig_logprob_cdf} that almost 90\% of the predicted log-probabilities of DirClip($-50$) posterior samples are smaller than its clipping value of $-50$. Similarly, almost 90\% of the predicted log-probabilities of DirClip($-10$) posterior samples are smaller than its clipping value of $-10$. Note that CIFAR-10 has 10 classes. Therefore, these models have converged approximately to predicting the clipping value for each wrong class and predicting a log-probability close to 0 for the true class.

For comparison, notice that cold posterior samples achieve perfect training accuracy (\cref{fig_dirclip_acc}) without predicting extremely small log-probabilities. This relates to the issue discussed in \Cref{sec_conf_prior}: the Dirichlet prior enforces small probabilities rather than directly enforcing high confidence (which is not the same thing). In particular, predicting extremely small log-probabilities negatively affects the test likelihood of the DirClip prior.

\subsection{Low likelihood}
\label{appendix_dirclip_low_likelihood}

In \Cref{fig_model_comparison}, we compare the accuracy and likelihood of various posterior distributions. Note that as we increase the concentration of the DirClip($-50$) prior, the train and test accuracy increase while the test likelihood \textit{decreases}. The opposite is true for the cold posterior: its test accuracy increases \textit{together} with the test likelihood.

\begin{figure}[t]
\begin{center}
\includegraphics[width=0.9\columnwidth]{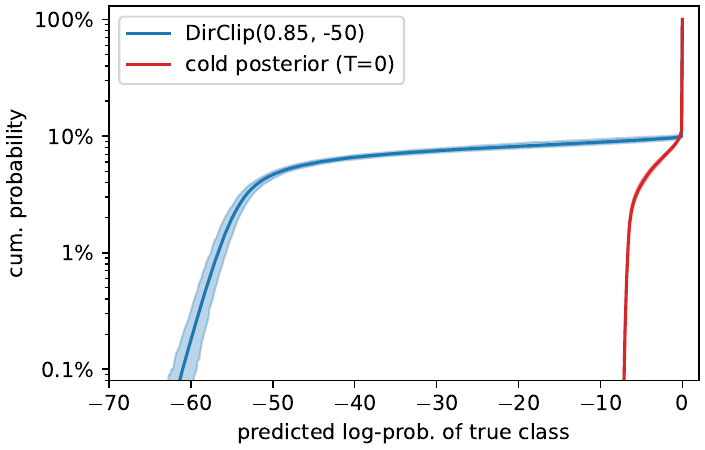}
\vskip -0.1in
\caption{\textbf{CDF of predicted log-probabilities for the true class on test data.} We sampled the DirClip and cold ResNet20 posteriors on CIFAR-10. Afterwards, we independently evaluated the predictions of each posterior sample on the test data, concatenating all predicted log-probabilities for the true class.}
\label{fig_cdf_logprob_true_class}
\end{center}
\end{figure}

The low likelihood of the DirClip($-50$) prior can be directly attributed to the small log-probabilities that it predicts. It not only predicts class log-probabilities of $-50$ on the training data (as shown in \Cref{fig_logprob_cdf}) but also on the test data, sometimes predicting a log-probability of $-50$ for the \textit{correct class}. Such a prediction incurs an extremely low likelihood. In \Cref{fig_cdf_logprob_true_class}, we show that the DirClip posterior samples misclassify almost $10\%$ of test images with log-probability of less than $-10$, while samples from the cold posterior have exactly \textit{zero} such overconfident misclassifications.

Intuitively, a BNN can misclassify an image with extremely high confidence only if all posterior samples assign an extremely low probability to the true label. Most of our experiments only used 8 (completely independent) posterior samples, so this happens often. If we instead used more posterior samples, it is more likely that at least one of the posterior samples would assign a higher probability to the true class, which would significantly improve the test likelihood of the DirClip prior. In \Cref{fig_posterior_size}, we show that as we increase the number of posterior samples, the likelihood of the DirClip prior indeed significantly improves. Whereas the likelihood of the cold posterior converges after only $\sim$20 posterior samples, the DirClip posterior requires 200 samples to reach a similar likelihood value.

On the one hand, this implies that the low likelihood of the DirClip posterior in \Cref{fig_model_comparison} is an artifact of our approximate inference pipeline, rather than a fundamental property of the true DirClip posterior distribution. On the other hand, the fact that the DirClip posterior requires an order of magnitude more posterior samples to converge (compared to a cold posterior) is a critical difference for any practitioner. We discuss the (high) computational cost of our experiments in \Cref{appendix_implementation_details}.

\subsection{fine-tuning has converged}

In \Cref{fig_dirclip_acc}, we compared the accuracy of both randomly initialized and fine-tuned DirClip models. Given that the fine-tuned models were initialized from a checkpoint with 100\% train accuracy, how can we know that the fine-tuning has converged? One check that we performed was comparing the CDF of predicted log-probabilities, as shown in \Cref{fig_logprob_cdf}. At initialization, the {\color{tab:orange}fine-tuned} DirClip model had a similar CDF to the {\color{tab:red}cold posterior}. However, after fine-tuning, the CDF converged to the CDF of the {\color{tab:blue}randomly initialized} DirClip model, suggesting that the fine-tuning has converged.

\section{Factorization of NDG posterior}
\label{appendix_ndg}

\begin{figure}[t]
\begin{center}
\includegraphics[width=0.7\columnwidth]{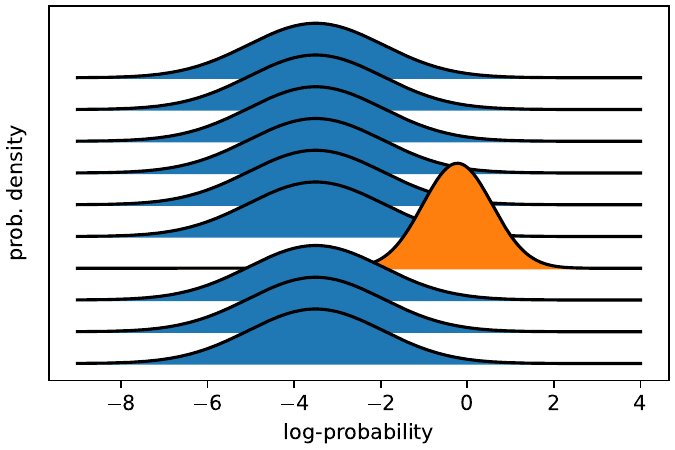}
\vskip -0.1in
\caption{\textbf{Visualization of the Noisy Dirichlet Gaussian (NDG) distribution over model's predicted log-probabilities.} The predicted log-probability for each class is assigned an independent Normal distribution. The log-probability corresponding to the {\color{tab:orange}correct} class has a shifted mean and reduced standard deviation; all {\color{tab:blue}other log-probabilities} have the same distribution.}
\label{fig_ndg_visualized}
\end{center}
\end{figure}

\citet{ndg} approximated the product of a Dirichlet$(\alpha)$ prior with the categorical likelihood using a Gaussian distribution over the model's predicted log-probabilities. They named this distribution \textit{Noisy Dirichlet Gaussian (NDG)}. We visualize the NDG distribution in \Cref{fig_ndg_visualized} and formally define its density function in \cref{eq_ndg_logpdf}.

\begin{align}
\begin{split}
\mathrm{NDG}(\log \mathbf{\hat{y}} | \boldsymbol{\tilde{\alpha}})_k &\sim \mathcal{N}(\log \hat{y}_k | \mu_k, \sigma_k^2), \ \textrm{where} \\
\tilde{\alpha}_k &= \alpha + \mathbb{1}(k=y) \\
\sigma_k &= \log (\tilde{\alpha}_k^{-1} + 1) \\
\mu_k &= \log(\tilde{\alpha}_k) - \log(\tilde{\alpha}_y) + (\sigma_y^2 - \sigma_k^2)/2
\label{eq_ndg_logpdf}
\raisetag{3\normalbaselineskip}
\end{split}
\end{align}

Technically, the original NDG model is a Gaussian distribution over predicted logits rather than log-probabilities. However, logits and log-probabilities are always equivalent up to a constant. Therefore, the only difference between defining the distribution over logits and log-probabilities is that using logits removes one degree of freedom. We experimentally verify that using NDG over logits and log-probabilities is equivalent in \Cref{fig_ndg_prior_likelihood_accuracy}.\footnote{When defining the distribution over log-probabilities, we shifted $\mu_k$ by a constant s.t.\ $\mu_y=0$.}

Unfortunately, it turns out that the NDG approximation effectively uses a quadratic likelihood term. Recall that NDG is not just an approximate prior; instead, it jointly approximates the prior \textit{together} with the likelihood. We factorize the $\mathrm{NDG}(\boldsymbol{\tilde{\alpha}})$ posterior in \cref{eq_ndg_factorization}, showing that it is equal to a combination of a reparameterized $\mathrm{NDG}(\boldsymbol{\alpha})$ prior and a quadratic likelihood term. The implied prior and likelihood functions are plotted in the left and middle plots of \Cref{fig_distribution_slices}.
\begin{align}
\begin{split}
\textrm{let } \log \hat{y}_k &\sim \mathcal{N}(\mu_k, \sigma_k^2) \ \textrm{where} \\
\mu_k &= \mu_1 \textrm{  if  } k=y \textrm{  else  } \mu_0 \\
\sigma_k &= \sigma_1 \textrm{  if  } k=y \textrm{  else  } \sigma_0 \\
\log p(\log \mathbf{\hat{y}}) &\stackrel{c}{=} -\frac{1}{2} \sum_k \frac{(\log\hat{y}_k - \mu_k)^2}{\sigma_k^2} \\
&\stackrel{c}{=} -\frac{1}{2} \left( \sum_k \frac{(\log\hat{y}_k - \mu_0)^2}{\sigma_0^2}\right) + \\
& \left( \frac{\sigma_0^2\mu_1-\sigma_1^2\mu_0}{\sigma_0^2 \sigma_1^2}\log\hat{y}_y + \frac{\sigma_1^2-\sigma_0^2}{2\sigma_0^2 \sigma_1^2}\log\hat{y}_y^2 \right)
\label{eq_ndg_factorization}
\raisetag{4.5\normalbaselineskip}
\end{split}
\end{align}

\section{Proof that confidence prior converges to a cold likelihood}
\label{appendix_confidence_prior}

In this section, we show that both a cold likelihood and the confidence prior combined with untempered categorical likelihood converge to a Dirac delta measure as $T \rightarrow 0$. We prove this by deriving lower and upper bounds on the product of the confidence prior with the categorical likelihood and showing that they both converge to the same distribution.

Recall from \cref{eq_conf_equals_cold_likelihood} that when the predicted class matches the true class, the density of the confidence prior combined with the categorical likelihood is equal to the density of a cold categorical likelihood. At the same time, the cold likelihood provides a lower bound on the product of the confidence prior with untempered categorical likelihood (because $\hat{y}_y \leq \max_k \hat{y}_k$). To obtain an upper bound on the product, note that conditional on a specific value of the predicted probability for the true class $\hat{y}_y$, the density of the confidence prior is maximized when the rest of the probability mass $(1-\hat{y}_y)$ is entirely concentrated in a single class. Combined, this yields the following lower and upper bounds:
\begin{equation}
\hat{y}_y^{1/T} \leq p(\boldsymbol{\hat{y}}) \cdot \hat{y}_y \leq \max (\hat{y}_y, 1-\hat{y}_y)^{1/T-1} \cdot \hat{y}_y.
\label{eq_confidence_bound}
\end{equation}

We show that both a cold likelihood and the confidence prior combined with the categorical likelihood converge to the Dirac delta measure by deriving the CDFs of both distributions and showing that the CDFs converge to the CDF of the Dirac delta measure as $T \rightarrow 0$.

We can think of the cold likelihood as an (unnormalized) distribution over the predicted probability of the true class. By doing so, we can derive its CDF:
\begin{align}
\begin{split}
F_\text{cold}(z) &= p(\hat{y} \leq z)\\
&= z^{1+1/T}.
\label{eq_cdf_cold}
\end{split}
\end{align}
Notice that as $T \rightarrow 0$, the cold likelihood CDF converges to the CDF of the Dirac delta measure concentrated at 1:
\begin{equation}
F_\delta(z) = 
\begin{cases}
0 & \text{$z < 1$} \\
1 & \text{$z = 1$}.
\end{cases}
\end{equation}

\begin{figure}[t]
\begin{center}
\includegraphics[width=0.7\columnwidth]{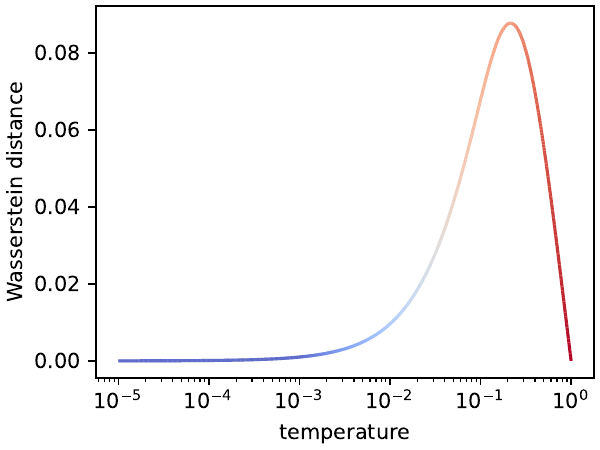}
\vskip -0.1in
\caption{Wasserstein distance between the cold likelihood and the upper bound on the product of confidence prior and untempered categorical likelihood.}
\label{fig_wasserstein}
\end{center}
\end{figure}

Similarly, we can derive the CDF for the upper bound on the product of the confidence prior and the untempered categorical likelihood introduced in \cref{eq_confidence_bound}:
\begin{align}
F_\text{up}(z) = 
\begin{cases}
\dfrac{2^{1/T}\left( T - (1-z)^{1/T}(T+z)\right)}{(2^{1/T}-1)(T+1)} & z \leq 0.5 \\
\\
\dfrac{2T+\frac{1}{1-2^{1/T}}}{2(T+1)} + \dfrac{(2z)^{1/T+1}-1}{2(2^{1/T}-1)(T+1)} \hspace{-2mm} &z > 0.5
\raisetag{3.2\normalbaselineskip}
\label{eq_cdf_conf}
\end{cases}
\end{align}
In the limit of $T \rightarrow 0$, the upper bound converges to the same distribution as the cold likelihood: the Dirac delta measure. At the same time, recall from \cref{eq_confidence_bound} that the cold likelihood provides a \textit{lower} bound for the product of the confidence prior and the categorical likelihood. Therefore, both the lower bound and the upper bound converge to the same distribution as the cold likelihood. In \Cref{fig_wasserstein}, we plot the Wasserstein distance between the cold likelihood and the upper bound. Notice that the Wasserstein distance is exactly zero at $T=1$ (because the confidence prior is equal to a uniform prior), it peaks around $T \approx 0.1$, and then again converges to zero as $T \rightarrow 0$.

\begin{figure}[t]
\begin{center}
\includegraphics[width=1\columnwidth]{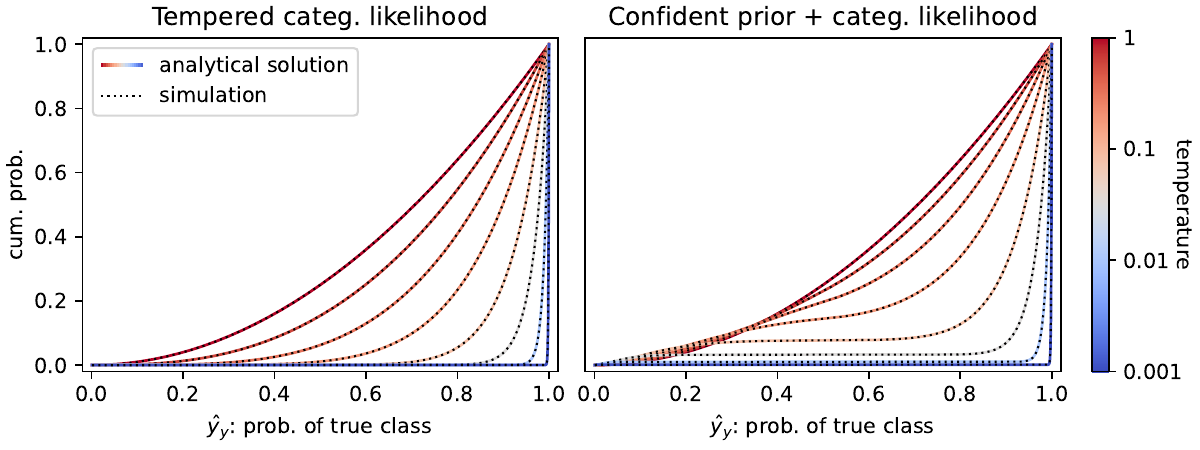}
\vskip -0.1in
\caption{\textbf{Left:} CDF of the categorical likelihood tempered to various temperatures (shown by the colorbar on the right). \textbf{Right:} CDF of the upper bound on the product of a confident prior (parameterized using temperature) with the untempered categorical likelihood. Notice that the analytical CDFs (solid lines) perfectly match the empirical CDFs obtained using Monte Carlo simulation (dotted lines).}
\vspace{-2mm}
\label{fig_cdf_validation}
\end{center}
\end{figure}

Lastly, to verify that the CDFs derived in \cref{eq_cdf_cold,eq_cdf_conf} are correct, we compared them to empirical CDFs obtained using Monte Carlo simulation in \Cref{fig_cdf_validation}.

\section{Priors over outputs}
\label{appendix_priors_over_outputs}

\subsection{Change of variables}
\label{appendix_change_of_variables}

Recall from \Cref{sec_dirichlet} that the Dirichlet prior is a distribution over model predictions $p(\mathbf{\hat{y}})$ whereas we ultimately want to sample a distribution over model parameters $p(\boldsymbol{\theta})$, as defined in \cref{eq_posterior}. So how can we translate the distribution over model predictions to a distribution over model parameters? Unfortunately, this is not as simple as setting $p(\boldsymbol{\theta})=p(\mathbf{\hat{y}})$. The issue is that probability \textit{density} is not conserved when transforming random variables; only probability \textit{mass} is conserved. Intuitively, the probability mass of $\boldsymbol{\theta}$ being in some small volume $|\Delta \boldsymbol{\theta}|$ is $p(\boldsymbol{\theta}) |\Delta \boldsymbol{\theta}|$, and by applying the same logic to $\mathbf{\hat{y}}$, we get:
\begin{align}
p(\boldsymbol{\theta}) |\Delta \boldsymbol{\theta}| &= p(\mathbf{\hat{y}}) |\Delta \mathbf{\hat{y}}| \\
p(\boldsymbol{\theta}) &= p(\mathbf{\hat{y}}) \frac{|\Delta \mathbf{\hat{y}}|}{|\Delta \boldsymbol{\theta}|}.
\end{align}
When the transformation is invertible, the change in volumes can be computed using the Jacobian determinant. Unfortunately in our case, the mapping from model parameters to model predictions is many-to-one, and therefore the standard Jacobian determinant method does not apply. For a further discussion of how this transform can be approximated, we recommend the work of \citet{fs_map}. We consider approximating this transformation to be beyond the scope of this paper, although it might be an exciting topic for future research.

We simply note that neither we nor \citet{ndg} have attempted to compute the ratio of volumes $\frac{|\Delta \mathbf{\hat{y}}|}{|\Delta \boldsymbol{\theta}|}$, i.e.\ the ``change of variables'' correction term. Without this correction term, setting the prior distribution over parameters to $p(\boldsymbol{\theta})=p(\mathbf{\hat{y}})$ does \textit{not} result in the prior predictive distribution following $\mathbf{\hat{y}} \sim p(\mathbf{\hat{y}})$. However, setting $p(\boldsymbol{\theta})=p(\mathbf{\hat{y}})$ does result in \textit{some} distribution over model parameters, and we simply need to ensure that this distribution is valid (i.e.\ integrable). When the Dirichlet prior is applied directly over model parameters, its integral diverges, so it is not a valid distribution (we prove this in \Cref{appendix_dirichlet_diverges}).

In contrast, the \textit{DirClip} prior density is bounded by some constant $p(\mathbf{\hat{y}}) \leq B$, so if the DirClip prior is combined with a proper prior over model parameters (e.g.\ $f(\boldsymbol{\theta}) = \mathcal{N}(0, 1)$), the integral of the joint density converges:
\begin{align}
\int p(\mathbf{\hat{y}}) f(\boldsymbol{\theta}) \mathrm{d} \boldsymbol{\theta} &\leq \int B f(\boldsymbol{\theta}) \mathrm{d} \boldsymbol{\theta} \\
&\leq B.
\end{align}
This implies that the DirClip prior combined with a vague Normal prior corresponds to a proper prior over model parameters.

\subsection{Proof that Dirichlet diverges}
\label{appendix_dirichlet_diverges}

For a probability density function to be valid, it must integrate to one. Technically, SGHMC doesn't require the distribution to be normalized, so the distribution only needs to integrate to a \textit{constant}. However, the integral of the Dirichlet prior over model parameters diverges. We prove this by looking at the probability density of a single model parameter, conditional on all other model parameters $p(\theta_1|\theta_2 \ldots \theta_N)$. More specifically, we took a ResNet20 trained on CIFAR-10 and fixed the values of all model parameters except a single ``bias'' parameter in the linear output layer, which we denote $\theta_1$.

\begin{figure}[t]
\begin{center}
\includegraphics[width=0.8\columnwidth]{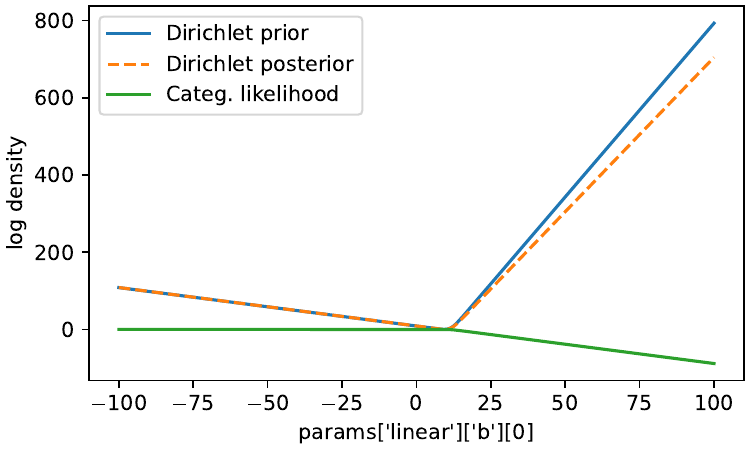}
\vskip -0.1in
\caption{\textbf{Dirichlet prior diverges}. We took a ResNet20 trained on CIFAR-10 and fixed the values of all parameters except a single bias parameter in the linear output layer. As we vary the value of this parameter, both the Dirichlet prior and posterior diverge.}
\vskip -0.15in
\label{fig_dirichlet_diverges}
\end{center}
\end{figure}

Recall from \cref{eq_dir_logpdf} that the density of the Dirichlet distribution is $\log p(\mathbf{\hat{y}}) \stackrel{c}{=} \sum_{k=1}^K (\alpha-1) \log \hat{y}_k$, which diverges to $\infty$ as one of the predicted probabilities $\hat{y}_k$ approaches zero. By changing the value of the neural network's bias parameter $\theta_1$, we are directly shifting the value of the output logit corresponding to the first class. Since the neural network is using the $\mathrm{softmax}$ activation function, changing the value of a single logit directly changes the predicted probabilities. As $\theta_1 \rightarrow \infty$, the predicted probability for the first class approaches $1$, while the predicted probability for all other classes approaches $0$, causing the Dirichlet probability density to diverge to $\infty$. Conversely, as $\theta_1 \rightarrow -\infty$, the predicted probability for the first class approaches $0$, thereby also causing the Dirichlet probability density to diverge to $\infty$. We numerically verified this using a ResNet20 trained on CIFAR-10, as shown in \Cref{fig_dirichlet_diverges}. However, we note that this behavior generalizes to any neural network using a linear output layer with a bias parameter and the $\mathrm{softmax}$ activation function.

In summary, the probability density of $p(\theta_1|\theta_2 \ldots \theta_N)$ diverges to $\infty$ as $\theta_1 \rightarrow \pm \infty$. Since the domain of $\theta_1$ is the real numbers, the integral $\int_\infty^{-\infty} p(\theta_1|\theta_2 \ldots \theta_N) \mathrm{d}\theta_1$ diverges, meaning that the conditional distribution of the first bias parameter is not a valid probability distribution. This implies that joint distribution over all model parameters is also not valid. Moreover, this divergence applies not only to the Dirichlet prior, but also the Dirichlet posterior that is obtained by combining the Dirichlet prior with a categorical likelihood. We show this is \Cref{fig_dirichlet_diverges}.

In practice, these are not just ``pedantic'' details. The fact that the Dirichlet posterior is not a valid distribution directly implies that it cannot be sampled. If we attempt to sample the Dirichlet posterior using SGHMC, or even optimize it using SGD, the model parameters will diverge. Both we and \citet{ndg} have observed this behavior.

\section{Dirichlet gradient step direction}
\label{appendix_dirichlet_gradient_direction}

In this section, we are interested in answering whether a small gradient step along the simplex of a probability distribution over classes will increase or decrease the probability of the true class. First, we derive a general formula for the update of any probability distribution. Second, we apply this formula to the Dirichlet distribution, discovering the update may decrease the probability of the true class when a small concentration parameter $\alpha$ is used.

Let's denote the predicted probabilities over classes $\mathbf{\hat{y}} = (\hat{y}_1, \hat{y}_2 \ldots \hat{y}_K)$. Since $\mathbf{\hat{y}}$ is only defined on the ($K$$-$$1$)-simplex (i.e.\ class probabilities must sum to 1), it is not possible to perform gradient ascent directly on $\mathbf{\hat{y}}$. Instead, it is more practical to parameterize the predictions using logits $\mathbf{z}$, where $\log \mathbf{\hat{y}} = \mathrm{logsoftmax}(\mathbf{z})$.

Additionally, let's denote the probability density function of interest $f$, so that the log density assigned to any prediction over classes is $\log f(\mathbf{\hat{y}}) = \log f(\mathrm{logsoftmax}(\mathbf{z}))$. A gradient ascent update of logits is therefore $\Delta \mathbf{z} = \epsilon \nabla_{\log f} \mathbf{J}_\mathrm{logsoftmax}$, where $\epsilon$ is the learning rate, $\nabla_{\log f}$ is the gradient of the log-PDF w.r.t.\ $\mathbf{\hat{y}}$ and the Jacobian of the logsoftmax function maps the change in $\mathbf{\hat{y}}$ to a change in $\mathbf{z}$.

Given a change in logits $\Delta \mathbf{z}$, we can linearly approximate the change in class log-probabilities $\Delta \log \mathbf{\hat{y}}$:
\begin{align}
\begin{split}
\Delta \log \mathbf{\hat{y}} &= \mathbf{J}_\mathrm{logsoftmax} \Delta \mathbf{z} \\
&= \epsilon \mathbf{J}_\mathrm{logsoftmax} \nabla_{\log f} \mathbf{J}_\mathrm{logsoftmax}.
\label{eq_class_prob_update}
\end{split}
\end{align}
To understand the update in \cref{eq_class_prob_update}, we must derive the Jacobian of the $\mathrm{logsoftmax}$ function:
\begin{align}
\mathrm{logsoftmax}(\mathbf{z})_i &= z_i - \log \sum_k \exp z_k \\
\frac{\partial \mathrm{logsoftmax}(\mathbf{z})_i}{\partial z_j} &= \mathbb{1}(i=j) - \hat{y}_j \\
\mathbf{J}_\mathrm{logsoftmax} &= I - \mathbf{1} \otimes \mathbf{\hat{y}},
\end{align}
and the gradient of the Dirichlet log-PDF:
\begin{align}
\log f(\mathbf{\hat{y}}) &= \sum_{k=1}^{K} (\alpha_k-1) \log \hat{y}_k \\
\nabla_{\log f} &= \boldsymbol{\alpha} - 1.
\end{align}
For convenience, let's denote the gradient of the log-PDF as $\mathbf{g} = \nabla_{\log f}$. Also, let's denote $g^+ = \sum_{k=1}^{K} g_k$ and $\hat{y}^{2+} = \sum_{k=1}^{K} \hat{y}^2_k$. Plugging this into \cref{eq_class_prob_update}, we get:
\begin{align}
\Delta \log \mathbf{\hat{y}} &= \epsilon \mathbf{J}_\mathrm{logsoftmax} (\nabla_\text{logpdf} \mathbf{J}_\mathrm{logsoftmax}) \\
&= \epsilon (I - \mathbf{1} \otimes \mathbf{\hat{y}}) (\mathbf{g} (I - \mathbf{1} \otimes \mathbf{\hat{y}})) \\
&= \epsilon (I - \mathbf{1} \otimes \mathbf{\hat{y}}) (\mathbf{g} - g^+ \mathbf{\hat{y}}) \\
&= \epsilon (\mathbf{g} - g^+\mathbf{\hat{y}} - \mathbf{\hat{y}} \cdot \mathbf{g} + g^+ \hat{y}^{2+}).
\end{align}
Gradient ascent increases the probability of the true class iff $\Delta \log \mathbf{\hat{y}}_y > 0$:
\begin{equation}
g_y - g^+\hat{y}_y - \mathbf{\hat{y}} \cdot \mathbf{g} + g^+ \hat{y}^{2+} > 0.
\label{eq_update_for_true_class}
\end{equation}
\cref{eq_update_for_true_class} provides a general condition that must hold for any distribution (and any particular prediction $\mathbf{\hat{y}}$) for the update of the true class to be positive. In \Cref{fig_dirichlet_phase_diagram}, we show how the update depends on both the distribution parameter $\alpha$ and the prediction $\mathbf{\hat{y}}$. To gain more insight into the Dirichlet distribution, we can look at the ``worst-case'' scenario in terms of $\mathbf{\hat{y}}$. In particular, we are interested in the case where the prior gradient dominates the likelihood gradient, and as a result, the combined gradient decreases the probability of the true class. Observe in \Cref{fig_dirichlet_gradient} that the gradient of the Dirichlet prior grows the closer the prediction gets to a single class. Therefore, the ``worst-case'' scenario is that $\mathbf{\hat{y}}$ is concentrated around a single class $j \ne y$. By applying this worst-case assumption, we get a minimum value of $\alpha$ that must be used so that the probability update for the true class is positive for all values of $\mathbf{\hat{y}}$:
\begin{align}
g_y - g^+\hat{y}_y - \hat{y}_j g_j + g^+ \hat{y}_j^2 &> 0 \\
g_y- g_j \hat{y}_j + g^+(\hat{y}_j^2 - \hat{y}_y) &> 0 \\
g_y- g_j + g^+ &\gtrsim 0 \\
\alpha - (\alpha-1) + (K\alpha - K + 1) &\gtrsim 0 \\
2 + K\alpha - K &\gtrsim 0 \\
K\alpha &\gtrsim K - 2 \\
\alpha &\gtrsim \frac{K - 2}{K} \label{eq_alpha_critical_value}
\end{align}

\begin{figure}[t]
\begin{center}
\includegraphics[width=0.9\columnwidth]{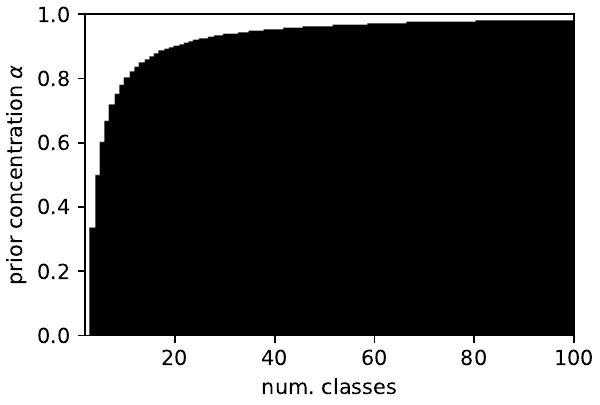}
\vskip -0.1in
\caption{\textbf{Phase diagram showing the ``critical value'' of the Dirichlet concentration parameter $\boldsymbol{\alpha}$.} When $\alpha$ is \textit{above} the critical threshold (i.e.\ in the white region), then a gradient step on the Dirichlet posterior will always increase the probability of the true class. On the other hand, when $\alpha$ is \textit{below} the critical threshold (i.e.\ in the black region), gradient steps might \textit{decrease} the probability of the true class, making optimization (or sampling) very challenging.}
\vskip -0.15in
\label{fig_dirichlet_critical_alpha}
\end{center}
\end{figure}

Notice that the critical value of $\alpha$ in \cref{eq_alpha_critical_value} depends on the number of classes $K$. This relationship is illustrated in \Cref{fig_dirichlet_critical_alpha}. For 10 classes, the critical value of $\alpha$ is 0.8, which is consistent with all of our experiments in \Cref{sec_dirclip}. As the number of classes grows, the critical value of $\alpha$ increases, approaching 1. Note that the increasing $\alpha$ \textit{decreases} the prior concentration; in particular, $\alpha=1$ corresponds to a \textit{uniform} prior. This is an unfortunate result, meaning that the Dirichlet prior gets increasingly unstable with a growing number of dimensions.

\section{Implementation details}
\label{appendix_implementation_details}

\textbf{Goal.}\quad In order to minimize bias when comparing different posterior distributions, we considered it important to obtain high-fidelity approximations for each posterior. In total, we spent approximately 30 \textit{million} epochs (750 TPU-core-days) to sample all ResNet20 posteriors on CIFAR-10. It was unfortunately this high computational cost that prevented us from testing more model architectures or datasets.

\textbf{Basic setup.}\quad All experiments were implemented in JAX \citep{jax} using TPU v3-8 devices. The typical way to draw SGHMC samples is to generate a single chain where each sample depends on the previous samples. Since BNN posteriors are multi-modal \citep{mode_connectivity, deep_ens, bdl}, generating autocorrelated samples means that the posterior distribution is explored slowly \citep{do_bnns_need_to_be_fully_stochastic}. To eliminate this autocorrelation (and thus achieve more accurate posterior approximations), we instead generated each posterior sample from a different random initialization, completely independently of the other samples. 

\begin{figure}[t]
\begin{center}
\includegraphics[width=0.7\columnwidth]{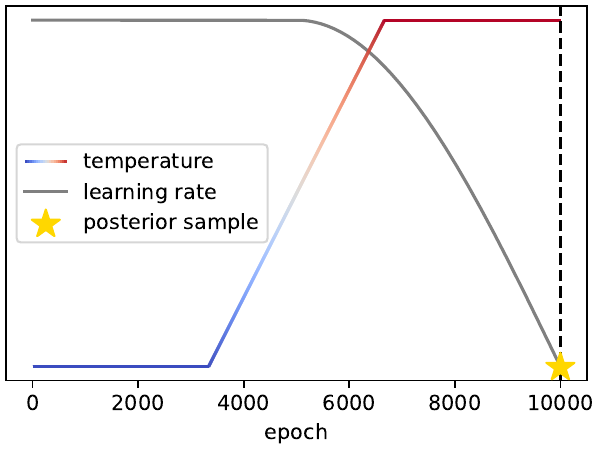}
\vskip -0.15in
\caption{\textbf{SGHMC learning rate and temperate schedule}. Temperature starts at zero, increases linearly after $\frac{1}{3}$ epochs and stays at the maximum value after $\frac{2}{3}$ epochs. Learning rate is constant for $\frac{1}{2}$ epochs and then follows a sine schedule to zero.}
\label{fig_sghmc_schedule}
\end{center}
\vskip -0.2in
\end{figure}

For most experiments, we followed the learning rate and temperature schedule depicted in \Cref{fig_sghmc_schedule}. Initially, the temperature is zero and the learning rate is high so that the sampler quickly converges to a local mode. Afterward, the temperature is increased to explore the posterior, and the learning rate is decreased to reduce the bias of the sampler. At the end of the cycle, only a single posterior sample is produced. We run this procedure in parallel across TPU cores to generate multiple posterior samples. We provide specific details for each experiment below.

\textbf{Training from scratch.}\quad For the experiments where we compared posterior tempering, DirClip, and NDG against each other (\cref{fig_ndg_prior_likelihood_accuracy,fig_dirclip_acc,fig_loss_landscapes_5x,fig_model_comparison,fig_logprob_cdf,fig_cdf_logprob_true_class}), a single posterior consists of 8 independent SGHMC samples, each using 10,000 epochs. To obtain error bars, this procedure was repeated three times, therefore leading to 24 posterior samples per model. However, these additional samples were \textit{only} used for uncertainty estimates; the mean accuracy and likelihood only correspond to a posterior consisting of 8 samples. Each posterior used a $\mathcal{N}(0, 0.1^2)$ prior over model parameters and a learning rate of $10^{-4}$. The learning rate was intentionally very low and the number of epochs very high so that each posterior sample would converge to the MCMC stationary distribution. 

\begin{figure*}[ht]
\begin{center}
\includegraphics[width=0.8\textwidth]{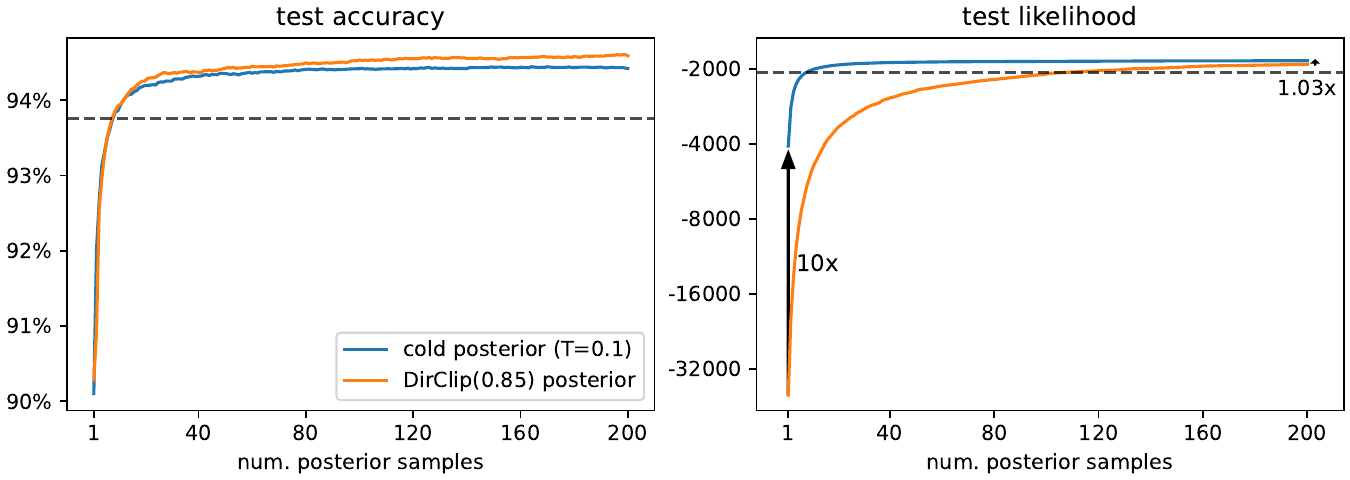}
\vskip -0.1in
\caption{\textbf{Posterior size}. Across all of our ResNet20 experiments, we used completely independent SGHMC posterior samples. The dashed line shows the test accuracy and likelihood for a cold posterior consisting of 8 samples. Observe that 8 samples are enough for the posterior accuracy of both a cold posterior and a DirClip posterior to converge within $1\%$. However, the test likelihood of the DirClip posterior takes much longer to converge.}
\label{fig_posterior_size}
\end{center}
\end{figure*}

\textbf{Are 8 posterior samples enough?}\quad Across all of our experiments, we used 8--16 \textit{independent} SGHMC samples to approximate each posterior distribution. This is a distinct approach to prior works which generated long correlated SGHMC chains. We chose this approach to eliminate any autocorrelation between posterior samples but also to obtain an algorithm that parallelizes more easily across TPU cores (allowing us to use relatively small batch sizes without incurring significant overhead). Using only 8 independent samples, and the same augmentation strategy as \citet{cold_posterior}, we matched or slightly exceeded the test accuracy of ResNet20 trained on CIFAR-10 of prior works \citep{cold_posterior,ndg,priors_revisited}. In \Cref{fig_posterior_size}, we show that we could further improve the test accuracy (and especially the test likelihood) by using more posterior samples. 

\textbf{Fine-tuning.}\quad In \Cref{fig_dirclip_acc,fig_model_comparison,fig_logprob_cdf}, some of the DirClip models were initialized from a pretrained model. More specifically, this means that the SGHMC sampler was initialized from a model with 100\% training accuracy, obtained using SGD. Given the unstable dynamics of fine-tuning, we ran the SGHMC sampler using a very low fixed learning rate and a fixed temperature $T=1$. The DirClip model with a clipping value of $-50$ used a learning rate of $10^{-7}$ and 10,000 epochs per sample. The DirClip models with a clipping value of $-10$ and $-20$ used an even lower learning rate $(3\cdot10^{-8})$ and an increased 50,000 epochs per sample, to ensure better numerical stability for small $\alpha$. 

\textbf{Confidence of a Normal prior.}\quad For the experiments shown in \Cref{fig_normal_conf,fig_temp_grid}, we sampled 15 different priors scales across 16 temperatures, leading to 240 unique posterior distributions. Given the large number of different distributions, we opted for the lowest sampling fidelity for this experiment. A single posterior consists of 16 independent SGHMC samples, each using 1,000 epochs. The learning rate was tuned for each prior scale to maximize the test accuracy of a single posterior sample. Once the learning rate was tuned, the 16 posterior samples were drawn using a different random seed, and the previous samples were discarded to avoid any data leakage. 

\textbf{2D toy example.}\quad \Cref{fig_dirclip_decision_boundary} shows the only experiment that we ran on a CPU rather than TPUs. We used a fully-connected neural network with 5 hidden layers of 10 neurons, using the ReLU activation function. We used HMC (\cref{appendix_hmc}) with 10,000 leapfrog steps and 1,500 samples per posterior. Both chains achieved an $\hat{R}$ metric \citep{r_hat} of 1.00 in function space and $\leq$1.05 in parameter space. We used a DirClip clipping value of $-50$.

\textbf{Training accuracy.} Note that in \Cref{fig_ndg_prior_likelihood_accuracy,fig_dirclip_acc}, the training accuracy was evaluated on posterior \textit{samples} rather than the posterior predictive distribution---the goal of these plots was to explicitly visualize the behavior of posterior samples. In contrast, test accuracy was measured using an ensemble of 8 posterior samples, to provide a direct measure of the posterior performance.

\textbf{Data augmentation.} We used exactly the same data augmentation strategy as \cite{cold_posterior}: left/right flip, border-pad, and random crop.

\section{Markov Chain Monte Carlo}
\label{appendix_mcmc}

In \Cref{sec_bnn}, we detailed a method to estimate the posterior predictive distribution of a BNN that requires access to samples from the posterior. Often these samples are generated using Markov-Chain Monte Carlo (MCMC), a family of algorithms used for sampling continuous distributions where direct sampling is not possible. MCMC generates a Markov chain of samples from the target distribution where each sample in the chain depends on the previous sample but is conditionally independent of all the samples that came before. The more dependent (correlated) these samples are, the lower the accuracy of our posterior predictive distribution approximation. 

Hamiltonian Monte Carlo (HMC) is often considered the ``gold standard'' of MCMC algorithms because it is capable of generating samples that have very low autocorrelation \citep{hmc}. \citet{bnn_posterior} have successfully sampled a ResNet20 posterior using HMC but this was at a cost of over 60 million epochs of compute. Due to the extreme computational cost, we instead relied on the cheaper SGHMC sampler, an approximate (and biased) method derived from HMC.

\subsection{Hamiltonian Monte Carlo}
\label{appendix_hmc}

The idea behind HMC is that instead of drawing samples directly from the target distribution $\pi(\boldsymbol{\theta})$, we define a new variable $\mathbf{v}$, called the \textit{momentum}, and draw samples from the joint distribution $\pi(\boldsymbol{\theta}, \mathbf{v}) := \pi(\boldsymbol{\theta}) \pi(\mathbf{v})$. Once we have the joint samples $(\boldsymbol{\theta}, \mathbf{v})$, we can discard the momentum $\mathbf{v}$ and we are left with samples from our target distribution $\pi(\boldsymbol{\theta})$.

HMC alternates between updating the momentum $\mathbf{v}$ on its own and jointly updating the parameters together with the momentum $(\boldsymbol{\theta}, \mathbf{v})$. In order to understand this process intuitively, it helps to think of $(\boldsymbol{\theta}, \mathbf{v})$ as the state of a physical particle rolling on a hill. The particle has a position $\boldsymbol{\theta}$, momentum $\mathbf{v}$, and the shape of this hill depends on the geometry of the target distribution. First, we flick the particle in a random direction to update its momentum. Then we let the particle roll for a few seconds, which jointly updates its position and momentum together.

More formally, when the momentum is updated on its own, it is drawn from a standard Normal distribution:
\begin{equation}
\mathbf{v} \sim \mathcal{N}(\mathbf{0}, I).
\end{equation}
The joint proposal for $(\boldsymbol{\theta}, \mathbf{v})$ is generated by simulating \textit{Hamiltonian dynamics}:
\begin{align}
\begin{split}
\mathrm{d} \boldsymbol{\theta} &= \mathbf{v} \mathrm{d}t \\
\mathrm{d} \mathbf{v} &= \nabla \log \pi(\boldsymbol{\theta}) \mathrm{d}t.
\end{split}
\end{align}
In theory, if the Hamiltonian dynamics were simulated exactly, the sampled values of $\boldsymbol{\theta}$ would follow the target distribution $\pi(\boldsymbol{\theta})$ exactly. Unfortunately, Hamiltonian dynamics describe a continuous system that we have to approximate in discrete steps, typically using the leapfrog algorithm. After each simulation of Hamiltonian dynamics, the new proposal for $\boldsymbol{\theta}$ will get accepted or rejected at random, with the acceptance rate depending on the accuracy of the numerical simulation. To achieve any reasonable acceptance rate (i.e.\ above $10\%$), the discretization steps need to be extremely small. However, each step requires computing the full-batch gradient of the posterior. For models with tens to hundreds of thousands of parameters, this might amount to more than $10,000$ steps (epochs) per proposal, or over 1 million epochs to generate 100 posterior samples. Naturally, this is extremely expensive and very rarely done in practice \citep{bnn_posterior}.

\vspace{-1.5mm}
\subsection{Stochastic Gradient Hamiltonian Monte Carlo}
\label{appendix_sghmc}

\begin{algorithm*}[ht]
\caption{SGHMC}
\label{alg_sghmc}
\begin{algorithmic}[1]
\State sample $\mathbf{v} \sim \mathcal{N}(\mathbf{0}, I)$ \Comment{sample momentum}
\For{$i \gets 1 \ldots n_\textrm{samples}$} \Comment{generate $n_\textrm{samples}$ from target distribution}
    \For{$i \gets 1 \ldots \textrm{B}$} \Comment{iterate over mini-batches}
        \State $\boldsymbol{\theta} \gets \boldsymbol{\theta} + \epsilon \mathbf{v}$ \Comment{position update}
        \State $\mathbf{v} \gets (1-\epsilon C) \mathbf{v} + \epsilon \nabla \log \tilde{\pi}(\boldsymbol{\theta}) + \mathcal{N}(\mathbf{0}, 2TCI\epsilon)$ \Comment{mini-batch momentum update}
    \EndFor
    \State $\boldsymbol{\theta}_i \gets \boldsymbol{\theta}$ \Comment{save new value}
\EndFor
\end{algorithmic}
\end{algorithm*}
\setlength{\dbltextfloatsep}{2.5mm}

In HMC, each update of momentum requires computing the full-batch gradient: $\Delta \mathbf{v} = \epsilon \nabla \log \pi(\boldsymbol{\theta})$. If we replace the full-batch gradient $\nabla \log \pi(\boldsymbol{\theta})$ with a mini-batch estimate $\nabla \log \tilde{\pi}(\boldsymbol{\theta})$, we can think of this as using the full-batch gradient but adding noise that arises from the mini-batch estimate: $\Delta \mathbf{v} = \epsilon \nabla \log \tilde{\pi}(\boldsymbol{\theta}) = \epsilon \nabla \log \pi(\boldsymbol{\theta}) + \mathcal{N}(\mathbf{0}, \epsilon^2 \mathbf{V})$. By the central limit theorem, the larger the mini-batch size is, the closer the noise approaches a Normal distribution. Denoting $\mathbf{B} := \frac{1}{2} \epsilon \mathbf{V}$, we can view the noisy mini-batch momentum update as a discretization of the following system:
\begin{align}
\begin{split}
\mathrm{d} \boldsymbol{\theta} &= \mathbf{v} \mathrm{d}t \\
\mathrm{d} \mathbf{v} &= \nabla \log \pi(\boldsymbol{\theta}) \mathrm{d}t + \mathcal{N}(\mathbf{0}, 2\mathbf{B}\mathrm{d}t).
\end{split}
\end{align}
Introducing noise to the dynamics biases the proposed samples toward a uniform distribution \citep{sghmc}. In theory, if we knew the exact scale of the noise, we could compensate for the noise exactly by introducing a friction term to the dynamics:
\begin{align}
\begin{split}
\mathrm{d} \boldsymbol{\theta} &= \mathbf{v} \mathrm{d}t \\
\mathrm{d} \mathbf{v} &= \nabla \log \pi(\boldsymbol{\theta}) \mathrm{d}t - B \mathbf{v} \mathrm{d}t + \mathcal{N}(\mathbf{0}, 2\mathbf{B}\mathrm{d}t).
\end{split}
\end{align}
Unfortunately, in practice, we don't know the true value of the noise scale, so we instead introduce a user-specified friction parameter $C$ to the system. This is where SGHMC turns from an exact method (with unbiased posterior samples) to an approximate method. During the momentum update, in addition to the stochastic mini-batch noise of scale $\mathbf{B}$, we inject Normal noise that exactly compensates for the friction:
\begin{align}
\begin{split}
\mathrm{d} \boldsymbol{\theta} &= \mathbf{v} \mathrm{d}t \\
\mathrm{d} \mathbf{v} &= \nabla \log \pi(\boldsymbol{\theta}) \mathrm{d}t - C \mathbf{v} \mathrm{d}t + \mathcal{N}(\mathbf{0}, 2CI\mathrm{d}t) + \mathcal{N}(\mathbf{0}, 2\mathbf{B}\mathrm{d}t).
\end{split}
\raisetag{2.25\normalbaselineskip}
\label{eq_sghmc_dynamics}
\end{align}
The friction no longer compensates for the combined noise and so the generated samples become biased. Fortunately, as $\epsilon \rightarrow 0$, the injected noise of scale $C$ dominates the mini-batch noise of scale $\mathbf{B} = \frac{1}{2} \epsilon \mathbf{V}$. Hence, as the step size decreases, the samples become asymptotically unbiased.

Another distinction to HMC is that, unlike standard Hamiltonian dynamics, the system described in \cref{eq_sghmc_dynamics} is not time-reversible, so computing an acceptance probability is not even possible. This further biases the generated posterior samples. At the same time, since there is no accept/reject step, it is possible to sample the momentum only once and keep it throughout the whole Markov chain, which increases the average trajectory length.

In the limiting behavior of $C \rightarrow 0$, SGHMC reduces to HMC although the posterior is approximated from mini-batches, there is no accept/reject step and the Hamiltonian dynamics are approximated using Euler's method instead of the leapfrog algorithm (leapfrog is more accurate \citep{hmc}). On the other end, if $C=1$, SGHMC reduces to Stochastic gradient Langevin dynamics \citep{sgld}.

Typically, a cold posterior is sampled by modifying the SGHMC algorithm instead of directly tempering the posterior. This is to avoid scaling the log-posterior, which could introduce numerical instability. More specifically, we scale the injected noise in line 5 of \cref{alg_sghmc} by the temperature $T$:
\begin{equation}
    \mathbf{v} \gets (1-\epsilon C) \mathbf{v} + \epsilon \nabla \log \tilde{\pi}(\boldsymbol{\theta}) + \mathcal{N}(\mathbf{0}, 2TCI\epsilon).
\label{eq_sghmc_with_temp}
\end{equation}
Naturally, when $T=1$, \cref{eq_sghmc_with_temp} is equivalent to the standard SGHMC algorithm but when $T<1$, it corresponds to sampling $\pi(\boldsymbol{\theta})^{1/T}$ with scaled hyperparameters $C$ and $\epsilon$ \citep{cold_posterior}.

%
%

\end{document}